\documentclass[11pt]{article}
\usepackage[final]{acl}
\usepackage{times}
\usepackage{latexsym}
\usepackage[T1]{fontenc}
\usepackage[utf8]{inputenc}
\usepackage{microtype}
\usepackage{inconsolata}
\usepackage{graphicx}
\usepackage{booktabs}
\usepackage{array}
\usepackage{tabularx}
\newcolumntype{Y}{>{\centering\arraybackslash}X}
\usepackage{multirow}
\usepackage{amsmath}
\usepackage{amssymb}
\usepackage{xcolor}
\usepackage{colortbl}
\usepackage{enumitem}
\usepackage{subcaption}
\usepackage{afterpage}

\definecolor{hred}{RGB}{168,26,38}
\definecolor{hblue}{RGB}{26,62,148}
\newcommand{\best}[1]{\textcolor{red}{\textbf{#1}}}
\newcommand{\second}[1]{\textcolor{blue}{\underline{#1}}}
\newcommand{\ourrow}{\rowcolor{yellow!15}}
\newcommand{\writerrow}{\rowcolor{black!7}}
\newcommand{\ours}{FrameCheck}
\newcommand{\one}{\mathbf{1}}
\definecolor{jwfill}{HTML}{FBD3B8}
\definecolor{icfill}{HTML}{CDE6F7}
\definecolor{delred}{RGB}{190,30,45}
\definecolor{insgreen}{RGB}{0,120,60}
\newcommand{\jwb}[1]{{\setlength{\fboxsep}{1pt}\colorbox{jwfill}{\strut #1}}}
\newcommand{\icb}[1]{{\setlength{\fboxsep}{1pt}\colorbox{icfill}{\strut #1}}}
\definecolor{spfill}{HTML}{D5EFD9}
\newcommand{\spb}[1]{{\setlength{\fboxsep}{1pt}\colorbox{spfill}{\strut #1}}}
\definecolor{szfill}{HTML}{E4DCF6}
\newcommand{\szb}[1]{{\setlength{\fboxsep}{1pt}\colorbox{szfill}{\strut #1}}}
\newcommand{\inp}[1]{{\ttfamily\footnotesize #1}}
\usepackage{soul}
\setstcolor{delred}
\newcommand{\deltw}[1]{{\color{delred}\st{#1}}}
\newlength{\dellen}
\newcommand{\delt}[1]{\settowidth{\dellen}{#1}{\color{delred}#1}\hspace{-\dellen}{\color{delred}\rule[0.55ex]{\dellen}{0.7pt}}}
\newcommand{\inst}[1]{\textcolor{insgreen}{\textbf{#1}}}

\title{Right Numbers, Wrong Story: When LLMs Narrate\\Financial Changes Beyond Their Inputs}

\author{%
  \textbf{Seunghan Lee, Jun Seo, Junhyeok Kang, Jaehoon Lee, Sangjun Han, Sungdong Yoo,} \\
  \textbf{Minjae Kim, Tae Yoon Lim, Dongwan Kang, Hwanil Choi, Soonyoung Lee, Wonbin Ahn} \\[4pt]
  LG AI Research
}

\begin{document}
\maketitle

\begin{abstract}
A financial report drafted by an LLM can state every number correctly but still go beyond its input, adding judgments and causes that no one supplied. We call these additions \textit{unsupported framing} and find them far more often in LLM drafts than in reports written by professionals, and attempts to fix them by prompting with how each word is used instead make such judgments more frequent. We therefore propose \textbf{\ours{}}, a plug-in that sits between the LLM and human review to keep each draft within its input. Specifically, \ours{} 1)~guides the draft with professional examples that report a change of similar magnitude, 2)~flags judgments that professionals would not use and causes the input does not give, and 3)~removes them by rule, without a model call. In experiments on 17 open LLMs and 3M drafts, \ours{} removes almost all unsupported framing from drafts and brings the judgment rate close to that of the writers of 10-K filings, the annual reports that US firms file with regulators. Code is available at \href{https://github.com/seunghan96/framecheck}{\texttt{github.com/seunghan96/framecheck}}.
\end{abstract}

\section{Introduction}
\label{sec:intro}

Large language models (LLMs) are increasingly applied to financial documents~\citep{wu2023bloomberggpt,islam2023financebench,lee2026summaries}, and to drafting annual reports and earnings releases. Current checks ask whether the numbers are right, by matching them across a report~\citep{hillebrand2022towards} or against filing data~\citep{hall2026verifin}. However, a draft can \textit{state every number correctly} and still \textit{tell a different story}, judging a change the data only report or explaining it with a cause no one gave.

\begin{figure}[t]
\centering
\includegraphics[width=\columnwidth]{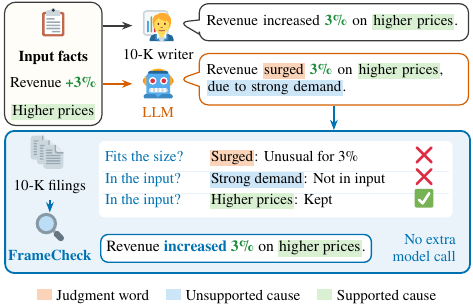}
\caption{\textbf{Same number, unsupported framing.} An LLM states the number correctly but \textit{adds a judgment word and an unsupported cause}. \ours{} checks words against 10-K filings and causes against the input.}
\label{fig:teaser}
\end{figure}

For example, a professional writer reported in a 10-K filing, the annual report of a US firm, that \textit{``Plywood average selling prices fell 22\% in 2000 compared to 1999.''} An LLM given only this change instead wrote that prices \textit{``decreased by 22.0\%, reflecting a \textbf{significant} decline in market demand and pricing pressures.''} Both sentences pass a fact check, yet only the draft adds a judgment (\textit{significant}), which the writer did not make, and a cause (\textit{market demand}) that its input never gave (Figure~\ref{fig:teaser}). Such \textit{unsupported framing} matters in regulated disclosure, as management must explain material changes in its own words and readers take the stated cause as a fact about the firm.

To see how often LLMs add such framing, we compare drafts of 17 open LLMs with the sentences that 10-K writers filed for the same changes. The drafts add magnitude words such as \textit{significantly} far more often than the writers, especially to large changes that writers report with the number alone. Most drafts also invent a cause for the change, often carrying the magnitude word inside it, as in \textit{``reflecting a significant decline''}. Furthermore, listing in the prompt when each word fits only increases such judgments, while asking the model to add nothing also removes the words that writers do use.

To this end, we propose \textbf{\ours{}}, a simple plug-in that sits between the LLM and human review to keep each draft within its input. Specifically, \ours{} guides each draft with 10-K sentences that report a change of similar magnitude, flags the judgments and causes that lack support, and removes them by rule. It checks each judgment word against its \textit{usage norm}, the range of change sizes on which 10-K writers use it, and also reports the \textit{judgment rate}, the share of texts that judge a change, as a measure of filing practice for each pipeline. Without training or any extra model call, \ours{} removes almost all unsupported framing from LLM drafts and brings their judgment rate close to the 1.5\% of the 10-K writers.

\noindent Our main contributions are summarized as follows:
\setlist[itemize]{leftmargin=0.3cm,itemsep=0pt,parsep=0pt,topsep=2pt,partopsep=0pt}
\begin{itemize}
\item We identify \textit{unsupported framing} in LLM-drafted financial reports, the judgments and causes that a draft adds beyond its input, and show that it changes how LLM readers judge the change.
\item We audit 17 open LLMs against 10-K writers and find that the LLM drafts add judgments far more often than the writers and give causes that the input lacks, while listing in the prompt when each word fits only increases such judgments.
\item We propose \ours{}, a model-free plug-in that keeps each draft within its input by guiding it with size-matched 10-K sentences, flagging unsupported judgments and causes, and removing them by rule without any model call.
\item Over 17 open LLMs, \ours{} cuts drafts with an unsupported cause from 77\% to 0.4\% and lowers the judgment rate from 17.2\% to 1.0\%, close to the 1.5\% of 10-K writers.
\end{itemize}

\section{Related Work}
\label{sec:related}

\noindent\textbf{Checking LLM-written financial text.}
LLMs are used to read, write, and summarize financial documents~\citep{wu2023bloomberggpt,chen2021finqa,islam2023financebench,kim2023bloated}. Checks for such text ask whether the stated facts are correct, by matching numbers across a report~\citep{hillebrand2022towards} or verifying each numeric statement against structured XBRL filing data~\citep{hall2026verifin,wang2026auditflow}. General faithfulness checks use trained classifiers~\citep{kryscinski2020evaluating,maynez2020faithfulness,laban2022summac,zha2023alignscore,tang2024minicheck}, atomic facts~\citep{min2023factscore}, chains of entailment checks~\citep{lei2023chain}, or LLM evaluators~\citep{zheng2023judging,liu2023geval,kim2024prometheus}, and post-hoc editors revise LLM output with self-feedback~\citep{madaan2023selfrefine}, verification questions~\citep{dhuliawala2024chain}, or a trained editor~\citep{mishra2024finegrained}, and style transfer replaces the words that mark an attribute~\citep{li2018delete}. These methods treat a sentence as correct once its stated facts are correct, overlooking the words that frame how large a change appears to the reader.

\noindent\textbf{Overstatement and verbal expressions.}
Prior work studies exaggeration in science communication~\citep{sumner2014association,wright2021semi,yu2019detecting}, the certainty of claims~\citep{pei2021measuring}, and hedges~\citep{farkas2010conll}, all learned from human labels. For LLMs, other work studies whether words of uncertainty are calibrated~\citep{lin2022teaching,mielke2022reducing,zhou2023navigating,belem2024perceptions} and measures the excess use of certain words in LLM-written text~\citep{kobak2025delving,liang2024mapping}. Unlike such word counts, our judgment rate is \textit{conditioned on the size} of the change, and our norms come from how professional writers use each word on changes of known size, not from human labels of sentences.

\noindent\textbf{Gradable words in disclosures.}
Gradable adjectives depend on a comparison class~\citep{kennedy2007vagueness,lassiter2017adjectival}, and NLP work ranks them by intensity~\citep{demelo2013good,kim2013deriving,gari2020bert} or grounds them and quantifiers in context and human judgments~\citep{sharp2018grounding,pezzelle2018some,li2023presque,tabach2026slightly}. Firms manage the tone and complexity of their disclosures~\citep{huang2014tone,bushee2018linguistic,cao2023machine}, measured with word lists from 10-K filings~\citep{loughran2011liability} or classifiers of statements in the Management's Discussion and Analysis (MD\&A)~\citep{li2010information}. We add the size of the change behind each magnitude word and use it to audit LLM-written disclosures.

\section{\ours{}: Grounding the Framing}
\label{sec:method}

\begin{figure*}[t]
\centering
\includegraphics[width=0.95\textwidth]{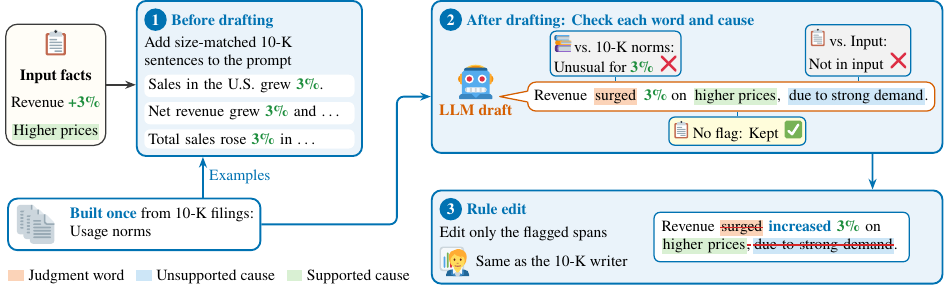}
\caption{\textbf{Overall framework of \ours{}.} Built once from 10-K filings of other firms, \ours{} 1)~adds size-matched 10-K sentences to the prompt, 2)~checks each judgment word against the usage norms and each cause against the input, and 3)~edits only the flagged spans by rule and without any model call, keeping the number.}
\label{fig:framework}
\end{figure*}

\begin{figure}[t]
\centering
\includegraphics[width=0.92\columnwidth]{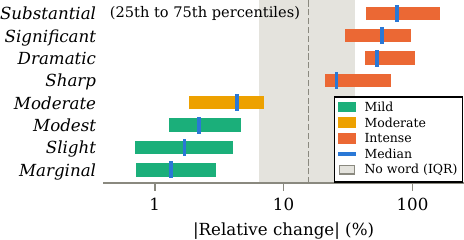}
\caption{\textbf{Magnitude words in 10-K filings.} Bars span the 25th to 75th percentiles of the change size for each word, against changes without a word (dashed: median).}
\label{fig:scale}
\end{figure}

\noindent\textbf{Overview.}
Figure~\ref{fig:framework} illustrates \ours{}. It builds usage norms from 10-K filings once and measures the judgment rate of the texts of a pipeline (Section~\ref{sec:norms}). For each draft, it then 1)~shows the model 10-K sentences on changes of a similar size, 2)~checks each judgment word against the norms and each cause against the input, and 3)~edits only the flagged texts by rule, with no model call. Working on text only, it plugs into any drafting pipeline, including those built on closed API models.

\subsection{Usage Norms and Judgment Rate}
\label{sec:norms}

We build usage norms from the MD\&A sections of 10-K filings in EDGAR-CORPUS~\citep{loukas2021edgar}, 220K filings from 38K firms filed from 1993 to 2020. The norms use only the \textit{reference firms} (70\%), while all evaluation items come from the \textit{test firms} (30\%). For each sentence that reports a change, we extract the change word (e.g., \textit{increased}), the magnitude word (e.g., \textit{slightly}), and the size of the change, from a stated percentage or two stated amounts. This yields 2.46M changes, 18K of them with a magnitude word (Appendix~\ref{app:extraction}).

\noindent\textbf{1) Usage norms.}
As shown in Figure~\ref{fig:scale}, the ranges of the words overlap, while their medians broadly rise from \textit{marginally} to \textit{substantially}, and most changes carry no word at all. We group the words into \textit{mild} (e.g., \textit{slightly}), \textit{moderate} (e.g., \textit{moderately}), and \textit{intense} (e.g., \textit{sharply}) classes. The usage norm of a word $w$ is the range from $q_{10}(w)$ to $q_{90}(w)$, the 10th and 90th percentiles of the change sizes on which writers use it (Appendix~\ref{app:norms}).

\noindent\textbf{2) Judgment rate.}
\ours{} finds magnitude words with a lexicon of 78 forms that covers adverbs (e.g., \textit{slightly}), adjectives (e.g., \textit{a modest increase}), and verbs (e.g., \textit{surged}). An adjective counts only when it modifies a change noun, negated words are skipped, and evaluative words such as \textit{strong}, which lack a size norm, are neither counted nor flagged. A single word is rarely wrong by itself, while a text that judges \textit{most} of its changes reads unlike a filing. \ours{} therefore reports the judgment rate by the size of the change and compares it with the 10-K writers' rate.

\subsection{Step 1: Size-Matched Retrieval}
\label{sec:align}

\noindent\textbf{Retrieval.}
Before drafting, \ours{} adds to the prompt six sentences of the reference firms that report a change in the same direction and of the closest size on a log scale. This follows example selection by similarity~\citep{liu2022makes}. As most such sentences carry no magnitude word, they show the model which word fits a size and how rarely a word is used. We retrieve only sentences without amounts or causes, which the model would otherwise copy into the draft (Appendix~\ref{app:retrieval}).

\noindent\textbf{Lightweight variants.}
For pipelines that cannot hold retrieved sentences, two lightweight variants replace retrieval with one sentence: \ours{} (style) states that 10-K filings usually report a change with its number alone and rarely add words such as \textit{slightly}, and \ours{} (faithful) asks to add nothing beyond the input. Each variant keeps Steps 2 and 3, unlike the prompt baselines of Section~\ref{sec:experiments}. \ours{} uses retrieval by default, and Section~\ref{sec:res_variant} compares all three variants.

\begin{table}[t]
\centering
\caption{\textbf{Two kinds of unsupported framing.} What each kind adds to a draft beyond its input.}
\label{tab:terms}
\small
\setlength{\tabcolsep}{3pt}
{\setlength{\aboverulesep}{0pt}\setlength{\belowrulesep}{0pt}\renewcommand{\arraystretch}{1.15}%
\begin{tabularx}{\columnwidth}{@{}l|X|X@{}}
\toprule
 & Adds & Checked against \\
\midrule
\multirow{2}{*}{\textbf{Judgment}} & How large it is & 10-K usage norms \\
\cmidrule(l){2-3}
 & \multicolumn{2}{l}{\textit{e.g.,} Revenue \textbf{\textit{surged}} 3\%.} \\
\midrule
\multirow{2}{*}{\textbf{Cause}} & Why it happened & Input of the draft \\
\cmidrule(l){2-3}
 & \multicolumn{2}{l}{\textit{e.g.,} Revenue rose 3\% \textbf{\textit{driven by demand}}.} \\
\bottomrule
\end{tabularx}}
\end{table}

\subsection{Step 2: Checking Words and Causes}
\label{sec:causes}

Step 2 asks whether each framing has support. As the input gives only the number, 1)~a judgment word needs support from the usage norms, while 2)~a cause needs support from the input (Table~\ref{tab:terms}).

\noindent\textbf{Judgment words.}
For a word $w$ on a change of size $x$, \ours{} flags an intense or moderate word as \textit{inflated} if $x < q_{10}(w)$, and a mild or moderate word as \textit{deflated} if $x > q_{90}(w)$. The flag is one-sided, as an intense word on a large change does not overstate it. As one in ten uses by the writers falls outside a range by definition, the flag marks a word for review rather than an error (Appendix~\ref{app:check}).

\noindent\textbf{Causes.}
A draft often explains the change with a cause that the input does not give (e.g., \textit{``driven by strong demand''}), and such a clause often carries a magnitude word itself (e.g., \textit{``reflecting a significant decline''}). \ours{} flags a clause that starts with a causal connective, such as \textit{due to}, \textit{driven by}, or \textit{reflecting}. When the input gives causes, as analyst notes do, the clause is kept if most of its content words occur in the input, with a threshold of 0.58 fit on 621 drafts of the reference firms. In a summary, a magnitude word absent from the source also counts as unsupported.

\subsection{Step 3: Editing Flagged Texts}
\label{sec:edit}

After drafting, a \textit{rule edit} cuts a flagged causal clause, deletes each flagged adverb or adjective (e.g., \textit{sharply}), and replaces a flagged verb (e.g., \textit{surged}) with \textit{increased} or \textit{decreased}, without a model call. Texts without a flag are kept, and the edit leaves the number and any supported cause in place. Appendix~\ref{app:edit} details the rules of this step.

\section{Experiments}
\label{sec:experiments}

\begin{table*}[t]
\centering
\caption{\textbf{Main results.} Shares of drafts (\%) over the 17 LLMs; $\approx$: closer to writers is better. Judg./Expl.: judgments and explanations by LLM judges with no word list. Misfit is 0 by design for \ours{}; Template is unranked.}
\label{tab:align}
\small
\setlength{\tabcolsep}{2.4pt}
\begin{tabular}{@{}l|cc|ccc|c|cc|c|c@{\hspace{\tabcolsep}}}
\toprule
 & \multicolumn{2}{c|}{\textbf{Judgment words}} & \multicolumn{3}{c|}{\textbf{Beyond the input}} & \multicolumn{1}{c|}{\textbf{Fidelity}} & \multicolumn{2}{c|}{\textbf{LLM judges}} & \multicolumn{1}{c|}{\textbf{Overall}} & \multicolumn{1}{c}{\textbf{Cost}} \\
\cmidrule(lr){2-3}\cmidrule(lr){4-6}\cmidrule(lr){7-7}\cmidrule(lr){8-9}\cmidrule(lr){10-10}\cmidrule(lr){11-11}
 & Judged $\approx$ & Misfit $\downarrow$ & Cause $\downarrow$ & Placeh. $\downarrow$ & Invented $\downarrow$ & Number $\uparrow$ & Judg. $\approx$ & Expl. $\downarrow$ & Gap $\downarrow$ & Calls $\downarrow$ \\
\midrule
Template & 0.0 & 0.00 & 0.6 & 0.0 & 0.0 & 100.0 & 0.2 & 0.0 & 2.1 & 0 \\
\midrule
Plain prompt & 17.2 & 0.35 & 77.0 & 8.5 & 0.9 & 98.0 & 19.4 & 79.2 & 101.2 & \best{0} \\
\midrule
\multicolumn{11}{l}{\textit{(a) \textbf{Prompt baselines}: Prompts that state the norms or ask for faithfulness}} \\
+ Norm card & 80.0 & 15.86 & 9.7 & \second{0.6} & \second{0.3} & 88.8 & 95.2 & 5.2 & 88.8 & \best{0} \\
+ Norm card with rate & 64.7 & 11.25 & 6.6 & 0.8 & 0.6 & 91.4 & 70.8 & 5.0 & 70.6 & \best{0} \\
+ Calibrated instruction & 20.7 & 1.81 & 6.5 & 1.0 & 1.6 & 99.0 & 22.6 & 6.4 & 26.7 & \best{0} \\
+ Style instruction & 0.1 & \best{0.00} & 5.3 & 3.1 & 2.3 & \best{99.8} & \second{0.2} & 5.2 & 9.8 & \best{0} \\
+ Faithful instruction & 0.3 & \best{0.00} & \second{0.6} & \second{0.6} & 0.6 & 99.4 & \second{0.2} & \best{0.2} & \second{2.3} & \best{0} \\
\midrule
\multicolumn{11}{l}{\textit{(b) \textbf{Sentence baselines}: Example sentences in the prompt}} \\
+ Balanced retrieval & 32.9 & 1.02 & 30.9 & 16.5 & 15.6 & 98.4 & 40.4 & 26.0 & 78.8 & \best{0} \\
+ Fixed sentences & 4.8 & \second{0.01} & 13.5 & 1.1 & 0.4 & \second{99.7} & 6.0 & 12.2 & 17.9 & \best{0} \\
+ Random-size sentences & 6.3 & 0.08 & 20.3 & 1.7 & 0.5 & \second{99.7} & 8.4 & 20.8 & 26.8 & \best{0} \\
\midrule
\multicolumn{11}{l}{\textit{(c) \textbf{Rule-based edit}: Deleting every magnitude word after drafting}} \\
+ Rule deleting all words & 0.0 & \best{0.00} & 77.0 & 8.5 & 0.9 & 98.0 & 4.4 & 79.2 & 87.0 & \best{0} \\
\midrule
\multicolumn{11}{l}{\textit{(d) \textbf{Post-hoc editors}: Models that revise the plain draft}} \\
DRG & \best{1.6} & 0.27 & 77.9 & 7.4 & 0.9 & 98.8 & 5.0 & 78.8 & 85.4 & \second{0.2} \\
Self-Refine & 9.2 & 0.39 & 49.3 & 10.2 & 5.6 & 98.2 & 10.8 & 54.4 & 67.2 & 3.3 \\
CoVe & 6.5 & 0.08 & 6.8 & \best{0.2} & \second{0.3} & 95.9 & 7.0 & 8.2 & 12.0 & 5.0 \\
FAVA & 16.9 & 0.35 & 75.7 & 8.5 & 0.9 & 98.0 & 19.2 & 77.6 & 99.6 & 1.0 \\
MiniCheck+rewrite & 6.2 & 0.06 & 6.3 & 0.7 & 1.0 & 99.3 & 6.4 & 8.0 & 11.7 & 1.8 \\
\midrule
\multicolumn{11}{l}{\textit{(e) \textbf{Ours}: Size-matched retrieval, the check, and the rule edit}} \\
\ourrow \ours{} & \second{1.0} & \best{0.00} & \best{0.4} & \best{0.2} & \best{0.1} & \best{99.8} & \best{1.4} & \second{0.4} & \best{1.1} & \best{0} \\
\specialrule{0.4pt}{1pt}{1pt}\specialrule{0.4pt}{0pt}{1pt}
\writerrow 10-K writers & 1.5 & 0.41 & -- & 0.0 & 0.0 & 75.1 & 1.8 & -- & -- & -- \\
\bottomrule
\end{tabular}
\end{table*}

\subsection{Experimental Setup}
\label{sec:setup}

\noindent\textbf{Models and tasks.}
We use 17 open instruction-tuned LLMs from six families (Qwen, Llama, Mistral, Gemma, Phi, and gpt-oss), from 7B to 72B parameters (Appendix~\ref{app:setup}). We test three separate tasks, with drafting as the main one. 1)~In \textit{drafting}, the model receives a line item, the fiscal year, and the signed change (e.g., \textit{Interest expense, fiscal 2001, +9.5\%}) and writes one MD\&A sentence, optionally with \textit{notes} that give the amounts and causes from the writer's sentence. 2)~In \textit{summarizing}, it shortens the writer's sentence, and 3)~in \textit{paragraph drafting}, it writes one MD\&A paragraph from a table of several changes reported in one filing (Appendix~\ref{app:dataset}). We take 980 changes from writers' sentences of the test firms, 70 for each of seven size bins and two directions, and each LLM writes four drafts per change. Drafting from the change alone isolates what a model adds beyond its input.

\noindent\textbf{Baselines.}
1)~\textbf{Prompt baselines} tell the model when writers use magnitude words: a \textit{norm card} lists the range of eight common words, optionally with the rate of writers who judge, a \textit{calibrated instruction} asks to judge only unusual changes, a \textit{style instruction} asks to follow 10-K filings, which rarely judge, and a \textit{faithful instruction} asks to add no cause or judgment beyond the input. 2)~\textbf{Sentence baselines} replace size-matched retrieval with six other 10-K sentences: three with and three without a magnitude word (\textit{balanced}), the same six for every change (\textit{fixed}), or six of \textit{random size}. 3)~A \textbf{rule-based edit} deletes every magnitude word, and a \textit{template} that only restates the input is shown as an unranked reference. 4)~\textbf{Post-hoc editors} revise the drafts: Delete-Retrieve-Generate (DRG)~\citep{li2018delete} with the words of a size-matched sentence, Self-Refine~\citep{madaan2023selfrefine}, CoVe~\citep{dhuliawala2024chain}, FAVA~\citep{mishra2024finegrained}, and MiniCheck+rewrite~\citep{tang2024minicheck}.

\afterpage{\begin{figure}[t]
\centering
\includegraphics[width=\columnwidth]{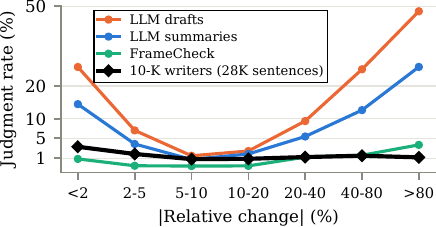}
\caption{\textbf{Judgment rate by size of change.} Share of sentences with a magnitude word over the 17 LLMs.}
\label{fig:register}
\end{figure}
\begin{table}[t]
\centering
\caption{\textbf{Judgment rate across tasks.} Texts with a judgment word (\%) over the 17 LLMs, best near the writers. Paragraphs count the words per 100 changes.}
\label{tab:robust}
\small
\setlength{\tabcolsep}{1.6pt}
\begin{tabular}{@{}l|cc|cc@{}}
\toprule
 & \multicolumn{2}{c|}{Drafting} & \multirow{2.5}{*}{Summary} & \multirow{2.5}{*}{Paragraph} \\
\cmidrule(lr){2-3}
 & w/o Note & w/ Note & & \\
\midrule
Plain prompt & 17.2 & 14.4 & 9.2 & 27.6 \\
Retrieval & 4.0 & 4.3 & 8.5 & 31.9 \\
\ourrow \ours{} & \best{1.0} & \best{1.6} & \best{0.3} & \second{18.4} \\
\ourrow \ours{} (style) & \second{0.0} & \second{0.0} & \second{0.0} & \best{2.1} \\
\specialrule{0.4pt}{1pt}{1pt}\specialrule{0.4pt}{0pt}{1pt}
\writerrow 10-K writers & 1.5 & 1.5 & 1.5 & 1.6 \\
\bottomrule
\end{tabular}
\end{table}
\begin{figure}[t]
\centering
\includegraphics[width=\columnwidth]{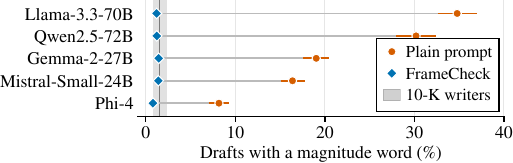}
\caption{\textbf{Judgment rate per drafting LLM.} The largest LLM of five families, with 95\% intervals.}
\label{fig:models}
\end{figure}
}
\noindent\textbf{Metrics.}
We measure drafts and writers with one lexicon on the same changes, and each metric is a share (\%). As Judged, Misfit, and Cause share the rules of \ours{}, three LLM judges without a word list recheck them (Appendices~\ref{app:metrics} and~\ref{app:ruleedit}).
\begin{itemize}[leftmargin=*,itemsep=0pt,topsep=2pt,parsep=0pt]
\item \textbf{Judgment words:} \textit{Judged} ($\approx$) is how often a text judges the change with a magnitude word, best when close to the writers, and \textit{Misfit} ($\downarrow$) is how often a word falls outside its usage norm.
\item \textbf{Beyond the input:} \textit{Cause}, \textit{Placeholder}, and \textit{Invented} ($\downarrow$) are how often a text adds a cause, a slot such as \textit{``[reason]''}, or an amount that is not given anywhere in the input of the draft.
\item \textbf{Fidelity:} \textit{Number} ($\uparrow$) is how often a text states the size of the change as given in the input.
\item \textbf{Overall:} \textit{Gap} ($\downarrow$) is the distance to the writers' judgment rate plus the shares with a cause or a placeholder, and \textit{Calls} counts the model calls per draft that a method adds after drafting. Writers' causes are not scored in Gap, as the writers knew causes that the drafting input does not give.
\end{itemize}

\subsection{Main Results}
\label{sec:res_audit}

\noindent\textbf{Comparison with baselines.}
Plain drafts judge 17.2\% of the changes, about 11 times the writers' 1.5\%, mostly with intense words, and 77.0\% add a cause \textit{that the input does not give} (Table~\ref{tab:align}). Prompts that list the sizes on which writers use each word raise the judgment rate, up to 80.0\%, even when they add that writers rarely judge, as LLMs often miss such negation~\citep{jang2023can}. Post-hoc editors cut few causes or still judge often, and DRG reaches the writers' rate but keeps causes in 77.9\%. The faithful instruction and a template without a model remove most framing (Gap 2.3 and 2.1), but they also drop the mild words that writers use, and the template cannot keep a cause that the input gives. \ours{} judges 1.0\% and leaves unsupported causes in 0.4\% of the drafts, with a Gap of 1.1, the lowest of all methods, and no model call after drafting. LLM judges with no word list agree (1.4\% against the writers' 1.8\%).

\noindent\textbf{Effect of the change size.}
LLM drafts judge 48.0\% of the changes above 80\% and summaries 26.6\%, whereas writers judge only 1.1\% and report such changes with the number alone (Figure~\ref{fig:register}). Real 10-K filings from 2016 to 2025 keep this practice after ChatGPT, also within firms ($-$0.14 points), so they remain a valid reference (Appendix~\ref{app:wild}).

\noindent\textbf{Robustness across tasks.}
With notes and in summaries, plain drafts still judge 14.4\% and 9.2\%, while \ours{} stays at 1.6\% and 0.3\% (Table~\ref{tab:robust}). Paragraphs carry 27.6 magnitude words per 100 changes against 1.6 of the writers, and retrieval even raises this, as models compare changes with other firms, so \ours{} reaches 18.4 and its style variant 2.1. Over three wordings, plain drafts judge 4.2\% to 17.2\%, and the mean Gap is 1.6 for \ours{}, 2.1 for the template, and 4.2 for the one-line faithful instruction (Appendix~\ref{app:prompts}).

\noindent\textbf{Consistency across models.}
Across the 17 LLMs (Figure~\ref{fig:models}), plain drafts judge 2.0\% to 35.1\% and \ours{} 0.0\% to 4.7\%, 0.5 points below the writers on the same changes (95\% CI $-1.3$ to 0.2).

\subsection{Analysis}
\label{sec:res_reader}\label{sec:res_fix}\label{sec:res_variant}

\noindent\textbf{Shift in LLM reader judgments.}
Eight open LLM readers rate versions of 600 changes with the same number for size, importance, and sentiment on 1 to 7 scales (Appendix~\ref{app:reader}). Inflating and deflating misfit words shift the rated size, readers take an invented cause as the reason for 99\% of the texts, and \ours{} removes almost all of the shift.

\begin{table}[t]
\centering
\caption{\textbf{Ablations of \ours{}.} Shares of drafts (\%) over the 17 LLMs. $^\ast$\ours{}, $^\dagger$\ours{} (style), and $^\ddagger$\ours{} (faithful).}
\label{tab:components}
\small
\setlength{\tabcolsep}{4.9pt}
\begin{tabular}{@{}l|ccc@{}}
\toprule
 & Judged $\approx$ & Misfit $\downarrow$ & Cause $\downarrow$ \\
\midrule
\multicolumn{4}{l}{\textit{(a) Adding the steps}} \\
Plain prompt & 17.2 & 0.35 & 77.0 \\
+ Step 1) Retrieval & 4.0 & 0.09 & \second{20.9} \\
+ Steps 2--3) Cause cut & \best{1.1} & \second{0.03} & \best{0.4} \\
\ourrow + Steps 2--3) Word edit$^\ast$ & \second{1.0} & \best{0.00} & \best{0.4} \\
\specialrule{0.4pt}{1pt}{1pt}\specialrule{0.4pt}{0pt}{1pt}
\multicolumn{4}{l}{\textit{(b) Adding Steps 2--3 to other prompts}} \\
Plain prompt & 17.2 & 0.35 & 77.0 \\
+ Cause cut \& Word edit & \best{8.2} & \best{0.00} & \best{2.8} \\
\midrule
Style instruction & \best{0.1} & \best{0.00} & 5.3 \\
\ourrow + Cause cut \& Word edit$^\dagger$ & 0.0 & \best{0.00} & \best{0.4} \\
\midrule
Faithful instruction & \best{0.3} & \best{0.00} & 0.6 \\
\ourrow + Cause cut \& Word edit$^\ddagger$ & \best{0.3} & \best{0.00} & \best{0.3} \\
\specialrule{0.4pt}{1pt}{1pt}\specialrule{0.4pt}{0pt}{1pt}
\writerrow 10-K writers & 1.5 & 0.41 & -- \\
\bottomrule
\end{tabular}
\end{table}

\noindent\textbf{Ablation study.}
Retrieval lowers the rate from 17.2\% to 4.0\% before drafting and the causes from 77.0\% to 20.9\%, the cause cut lowers them to 1.1\% and 0.4\% by removing invented causes with the words inside them, and the word edit removes the last misfit words (Table~\ref{tab:components}). In the edit, a fixed 10\% threshold works as well as the norms, which matter most for measuring. After other prompts, Steps 2 and 3 also cut causes to 2.8\% or less, but only retrieval keeps the judgment rate near the writers. Cutting causes alone halves the judged plain drafts, as many words sit in causes, and deleting every word also drops writers' mild words (Appendix~\ref{app:stats}).

\begin{table}[t]
\centering
\caption{\textbf{Drafting from a table and a note.} Six LLMs on 300 tables and notes. Words per 100 changes; causes beyond and supported by the note, by LLM judges.}
\label{tab:real}
\small
\setlength{\tabcolsep}{1.2pt}
\begin{tabular}{@{}l|ccc@{}}
\toprule
 & Words $\approx$ & Unsupported $\downarrow$ & Supported $\uparrow$ \\
\midrule
Plain prompt & 32.1 & 49.0 & 73.3 \\
\midrule
+ Retrieval & 33.9 & 52.0 & 71.4 \\
\ourrow \ours{} & \best{20.1} & \best{24.5} & \best{82.5} \\
\midrule
+ Style & 4.8 & 29.5 & 81.3 \\
\ourrow \ours{} (style) & \best{3.5} & \best{20.5} & \best{86.6} \\
\midrule
+ Faithful & 7.2 & 10.0 & 91.9 \\
\ourrow \ours{} (faithful) & \best{5.6} & \best{8.0} & \best{93.3} \\
\specialrule{0.4pt}{1pt}{1pt}\specialrule{0.4pt}{0pt}{1pt}
\writerrow 10-K writers & 3.4 & 28.9 & 83.5 \\
\bottomrule
\end{tabular}
\end{table}

\noindent\textbf{Variants for other inputs.}
With notes, the support test keeps most supported causes, while cutting every cause drops nearly all (Appendix~\ref{app:cause}). Closer to deployment, each of six LLMs drafts a paragraph from a table of changes and a note of the writer's causes, and plain drafts still add causes beyond the note in 49.0\%. \ours{} (style) cuts the words from 32.1 to 3.5 per 100 changes and these causes to 20.5\%, against the writers' 3.4 and 28.9\% (Table~\ref{tab:real}), while keeping most noted causes. The faithful instruction leaves fewer such causes (10.0\%) but states the number less often, and \ours{} (faithful) cuts its words and such causes further (5.6 and 8.0\%). We suggest retrieval for sentences and the prompt variants for paragraphs.

\noindent\textbf{Other analysis.}
Appendices~\ref{app:writers} to~\ref{app:examples} also cover writers on good and bad news, the cause check, all 17 LLMs, cost, detection, prior editors given our feedback, other norms, and forced word choice.

\section{Conclusion}
\label{sec:conclusion}

We proposed \ours{}, a plug-in that keeps LLM-written financial text within its input, with norms from 10-K filings. LLM drafts judge far more often than 10-K writers, often inside unsupported causes, though their words mostly fit the size. \ours{} closes most of this gap for single sentences and its style variant for paragraphs, while listing word ranges backfires. A correct number does not make a correct sentence.

\section*{Limitations}

Our norms describe how 10-K writers use magnitude words, not how readers read them. Words such as \textit{material} carry a legal sense, which a skip rule only partly separates (Appendix~\ref{app:cause}).

Our cause check relies on a list of connectives and misses causes introduced by other verbs or placed in the subject (Appendix~\ref{app:check}). Judged, Misfit, and Cause share this list and the lexicon of \ours{}, which we offset with LLM judges that use no word list. Item 303 of Regulation S-K asks MD\&A to explain material changes, and a cause cut by \ours{} therefore marks a place where the preparer should supply the true cause. The main drafting task gives no cause by design, so any cause there is unsupported, while with notes, as in practice, \ours{} keeps almost all of the causes that these notes give to the model.

All readers and gold labels come from LLMs, some of which also draft, and a study with finance professionals remains future work. The default \ours{} also leaves many more judgment words in paragraphs than its style variant.

All norms come from US 10-K filings in English, and earnings releases, other markets, and other languages may use the words differently. The extraction makes errors on a small share of pairs, and the check assumes one change per text, which causes false flags for summaries and paragraphs that report several changes at once.

Open LLMs may have seen some 10-K filings during pretraining, a known source of look-ahead bias in financial NLP~\citep{lee2026palm}. We evaluate open models up to 72B parameters, including the open-weight gpt-oss of OpenAI, and no closed models, although \ours{} does not need access to the model. Finally, matching the practice of 10-K writers is a choice of style, and other uses, such as investor letters, may call for a different one.

\section*{Ethical Considerations}

All data are public filings of the US SEC and public news corpora. The norms describe the past practice of 10-K writers, which may include strategic wording, such as the understatement of bad news. Aligning LLM drafts with these norms should not be read as an endorsement of such practice, and \ours{} therefore flags both inflated and deflated words, not only overstatement. \ours{} is meant to support human review of financial text, not to replace the people who sign it.

\bibliography{references}

\clearpage
\appendix
\setcounter{dbltopnumber}{3}\setcounter{topnumber}{3}\setcounter{totalnumber}{6}\renewcommand{\topfraction}{0.9}
\section{Anticipated Questions}
\definecolor{qared}{HTML}{A31515}
\newcommand{\qaref}[1]{{\hypersetup{linkcolor=qared}\textcolor{qared}{#1}}}

\label{app:qa}

This section answers questions that readers may raise, each with a pointer to the evidence for its answer in the main text or in the appendix.

\noindent\textbf{Q1. Is a magnitude word that fits its norm an error?}
No. A single word is rarely wrong by itself, so Step 2 flags only words outside their usage norm. The judgment rate measures a pipeline instead, as a pipeline whose texts judge most changes reads unlike a filing even when each word fits \qaref{(Section~\ref{sec:norms})}.

\noindent\textbf{Q2. Are the main metrics circular with the method?}
Partly. Judged, Misfit, and Cause share the lexicon and connectives of \ours{}, and Misfit is therefore 0 by design. Three LLM judges with no word list recheck every method in the Judg.\ and Expl.\ columns of Table~\ref{tab:align}, with Fleiss $\kappa$ of 0.96 and 0.94. They place \ours{} closest to the writers' 1.8\% on Judg., while the faithful instruction leaves fewer explanations (0.2\% vs.\ 0.4\%). Three other judges, two of which wrote none of the drafts, confirm that the rule edit removes causes and keeps complete sentences \qaref{(Appendix~\ref{app:ruleedit})}.

\noindent\textbf{Q3. Why match the writers' rate rather than zero?}
Writers do judge 1.5\% of these changes, and we take their practice as the target. With a target of zero, the template (0.6) and the faithful instruction (1.5) rank ahead of \ours{} (1.6). We therefore also report each of the three parts of the Gap separately \qaref{(Table~\ref{tab:align}, Appendix~\ref{app:metrics})}.

\noindent\textbf{Q4. How much does \ours{} gain over a template or a faithful instruction?}
On single sentences, about one point of Gap (Table~\ref{tab:align}). The template needs no model but cannot write paragraphs or use notes. Steps 2 and 3 also apply after the faithful instruction, which gives \ours{} (faithful), and lower its words and causes further in paragraphs from a table and a note \qaref{(Tables~\ref{tab:components} and~\ref{tab:real})}.

\noindent\textbf{Q5. Is cutting every cause not trivially best when the input has none?}
In the main task, yes, by design, to isolate what a model adds. With notes, the support test of Step 2 keeps most supported causes, and in paragraphs from a table and a note, every variant of \ours{} keeps most of the noted causes \qaref{(Appendices~\ref{app:para} and~\ref{app:cause})}.

\noindent\textbf{Q6. Do the usage norms matter if a fixed threshold works in the edit?}
Mainly for measurement and review. Plain drafts rarely misfit, so inside the edit a fixed 10\% threshold works as well (Section~\ref{sec:res_fix}). The norms set the writers' rate by size, catch the misfit words of the norm card (15.9\%), and beat a fixed threshold on moderate words and on words injected into drafts \qaref{(Appendix~\ref{app:detect})}.

\noindent\textbf{Q7. Do the results hold for each model?}
The norm card raises the rate for every model, and retrieval lowers or keeps it for 15 of the 17 models. Single-model intervals are narrower than the pooled ones, and most of them exclude the writers' own rate of 1.5\% \qaref{(Appendices~\ref{app:wm} and~\ref{app:stats})}.

\noindent\textbf{Q8. Why the 10th to 90th percentiles?}
As a common default. A wider range flags fewer of the writers' own sentences with the same recall, a narrower one flags many of them, and norms from other filings flag almost the same drafts. Plain drafts rarely misfit under every range \qaref{(Appendix~\ref{app:sensitivity})}.

\noindent\textbf{Q9. Why does the default variant do worse on paragraphs?}
Retrieved sentences lead models to compare changes with other firms. We therefore suggest the style or faithful variant, each adding one sentence to the prompt, for paragraphs of several changes \qaref{(Tables~\ref{tab:robust} and~\ref{tab:real}, Appendix~\ref{app:deploy})}.

\noindent\textbf{Q10. Do the results depend on the instruction P0?}
The rates do, as plain drafts judge 4.2\% to 17.2\% over three wordings and \ours{} judges 0.2\% under P1 and P2. The ranking holds over all three wordings, with a mean Gap of 1.6 for \ours{}, 2.1 for the template, and 4.2 for the faithful instruction \qaref{(Section~\ref{sec:res_audit}, Appendix~\ref{app:prompts})}.

\noindent\textbf{Q11. Does cutting a cause conflict with Item 303 of Regulation S-K?}
Item 303 asks MD\&A to explain material changes. The slot mode therefore replaces a cut cause with \textit{``[explain driver]''}, so that the preparer supplies the true cause, and a cause that the notes support is kept \qaref{(Appendix~\ref{app:deploy})}.

\noindent\textbf{Q12. Why are there no human raters or closed models?}
All readers and gold labels come from open LLMs rather than people, and a study with finance professionals remains future work. \ours{} works on text only and therefore needs no access to the model, while the rates of closed models remain unmeasured \qaref{(Limitations)}.

\noindent\textbf{Q13. Are the test changes contaminated?}
Norms and retrieval use only the reference firms. Of the 980 test sentences, 6\% have an exact copy in the pool, and removing near-duplicates barely changes the rate. The retrieval restriction, set after seeing test drafts, also holds on development changes, and in a memorization test no completion reproduces a writer's own sentence \qaref{(Appendix~\ref{app:retdesign})}.

\noindent\textbf{Q14. What does \ours{} cost?}
No model call after drafting, against up to five for prior editors (Table~\ref{tab:align}). Steps 2 and 3 take about 0.04 seconds per 1{,}000 drafts on a CPU, and retrieval adds 173 prompt tokens per draft \qaref{(Appendices~\ref{app:cost} and~\ref{app:deploy})}.

\clearpage
\section{Experimental Details}
\label{app:setup}

\subsection{Dataset}
\label{app:dataset}

\noindent\textbf{10-K filings.}
We use the MD\&A sections of 10-K filings in EDGAR-CORPUS~\citep{loukas2021edgar}, 220K filings from 38K firms filed from 1993 to 2020. The 20.9K firms with an extracted change are split at random into reference firms (70\%), which build the usage norms and the retrieval pool, and test firms (30\%), which give every evaluation item (Appendix~\ref{app:norms}). For recent filings, we also add the 10-K filings from 2021 to 2025 from a public extension of the same EDGAR corpus (Appendix~\ref{app:wild}).

\noindent\textbf{Drafting items.}
We sample 980 changes of the test firms, 70 for each of seven size bins and two directions, each from a sentence with a plain line item (e.g., revenue). The writer's own sentence for each change is the reference of that change. Table~\ref{tab:tasks} compares the three tasks and their inputs.

\begin{table*}[t]
\centering
\caption{\textbf{Three evaluation tasks.} Each task runs separately on its own input, and no output feeds another task.}
\label{tab:tasks}
\small
\begin{tabularx}{\textwidth}{@{}l|>{\raggedright\arraybackslash}X|l@{}}
\toprule
Task & Input to the LLM & Output \\
\midrule
1) Drafting (main) & Line item, fiscal year, and signed change & One MD\&A sentence \\
\quad with notes & The same, with the amounts and causes of the writer's sentence & One MD\&A sentence \\
2) Summarizing & The writer's own 10-K sentence, not an LLM draft & One summary of at most 15 words \\
3) Paragraph drafting & A table of four changes from one filing & One MD\&A paragraph \\
\bottomrule
\end{tabularx}
\end{table*}

\noindent\textbf{Other inputs.}
With \textit{notes}, an item also carries the amounts and causes from the writer's sentence with its magnitude words removed. A \textit{summary} item gives the writer's sentence, and a \textit{paragraph} item is a table of four changes from a test-firm filing. The realistic setting of Section~\ref{sec:res_variant} uses 300 more paragraphs with three to five changes, as the writer reported them, and a note of the writer's causes. Table~\ref{tab:task_examples} shows each task on test changes.

\begin{table*}[t]
\centering
\caption{\textbf{Examples of the three tasks.} Input to the LLM and the plain output of Llama-3.3-70B. Judgment words (\jwb{orange}), causes the input does not give (\icb{blue}), and causes it gives (\spb{green}) are marked by \ours{}.}
\label{tab:task_examples}
\small
\begin{tabular}{@{}>{\raggedright\arraybackslash}p{2.0cm}|>{\raggedright\arraybackslash}p{6.2cm}|>{\raggedright\arraybackslash}p{6.5cm}@{}}
\toprule
Task & Input to the LLM & Plain LLM output \\
\midrule
Drafting & \inp{Line item: Freight and mail revenues\newline Period: fiscal 1996 compared with fiscal 1995\newline Change: -1.7\%} & Freight and mail revenues decreased by 1.7\% in fiscal 1996 compared to fiscal 1995, reflecting a \jwb{modest} \icb{decline in shipping} \icb{volumes and revenue yields}. \\
\midrule
Drafting with notes & The same input, after the line \inp{Notes from the finance team: Freight and mail revenues decreased 1.7\% reflecting increased competition in the Alaska markets.} & Freight and mail revenues decreased by 1.7\% in fiscal 1996 compared to fiscal 1995, primarily due to \spb{increased competition} \spb{in the Alaska markets}. \\
\midrule
Summarizing & The writer's sentence: \textit{Freight and mail revenues decreased 1.7\% reflecting increased competition in the Alaska markets.} & Freight and mail revenues down 1.7\% due to \spb{Alaska competition}. \\
\midrule
Paragraph from a table & \inp{Changes in line items, fiscal 2002 compared with fiscal 2001:\newline - Office revenues: -4.0\%\newline - Other revenues: -9.0\%\newline - DMO revenue: -13.0\%\newline - Production revenue: -7.0\%} & Our revenues for fiscal 2002 declined compared to fiscal 2001, driven by decreases across multiple revenue streams. Office revenues decreased by 4.0\% year-over-year, while other revenues fell by 9.0\%. The decline in DMO revenue was more pronounced, with a 13.0\% decrease from fiscal 2001. Additionally, production revenue declined by 7.0\% during the same period. These decreases reflect \icb{a challenging operating} \icb{environment and competitive} \icb{market conditions} that impacted our ability to generate revenue growth in fiscal 2002. \\
\bottomrule
\end{tabular}
\end{table*}

\subsection{Backbone}
\label{app:backbone}

\noindent\textbf{Drafting LLMs.}
The drafting LLMs are Qwen2.5-7B, 14B, 32B, and 72B~\citep{qwen2025qwen25}, Qwen3-8B, 14B, and 32B~\citep{yang2025qwen3}, Llama-3.1-8B, Llama-3.1-70B, and Llama-3.3-70B~\citep{grattafiori2024llama3}, Mistral-7B~\citep{jiang2023mistral}, Mistral-Small-24B~\citep{mistral2025small}, Gemma-2-9B and 27B~\citep{gemma2024gemma2}, Gemma-3-12B~\citep{gemma2025gemma3}, Phi-4~\citep{abdin2024phi4}, and gpt-oss-20B~\citep{openai2025gptoss}. Mistral-Small-24B, the 32B models, and the 70B and 72B models are 4-bit AWQ checkpoints, Qwen3 runs without its thinking mode, and gpt-oss uses low reasoning effort. The \textit{10-model subset}, all but Qwen2.5-32B, Qwen3-14B, Qwen3-32B, Llama-3.3-70B, Mistral-Small-24B, Gemma-2-27B, and gpt-oss-20B, serves the judged samples, the prior editors given our feedback, and detection on real drafts. All other results use the 17 LLMs, except the paragraphs from a table and a note (six LLMs) and the reader study (eight LLMs). Over all conditions, the LLMs wrote about 3.1M drafts, four per change, condition, and model, and each main result uses 17 LLMs $\times$ 980 changes $\times$ 4 drafts.

\noindent\textbf{Other models.}
The judges are Llama-3.3-70B, Mistral-Small-24B, Qwen2.5-32B, Qwen2.5-72B, Qwen3-32B, and gpt-oss-20B, and the reader study uses eight open LLMs from 7B to 72B parameters (Appendix~\ref{app:reader}). MiniCheck-FT5 and Bespoke-MiniCheck-7B~\citep{tang2024minicheck} serve as support checkers and baselines, and all models are open checkpoints used with no fine-tuning.

\subsection{Decoding, Prompts, and Baselines}
\label{app:expsetup}

\noindent\textbf{Decoding.}
We serve all models with vLLM~\citep{kwon2023efficient} on NVIDIA L40S GPUs. Each model writes four samples per change at temperature 0.7 and top-$p$ 0.95 with a fixed seed, and the LLM judges decode greedily with a single output.

\noindent\textbf{Instructions.}
A drafting input looks like \textit{Interest expense, fiscal 2001 compared with fiscal 2000, +9.5\%}, under three paraphrases of the drafting instruction (P0 to P2). A summary has at most 15 words, and Appendix~\ref{app:prompts} gives all prompts.

\noindent\textbf{Baselines.}
Each prompt baseline adds one text to the prompt (Appendix~\ref{app:prompts}). Each sentence baseline adds six 10-K sentences of the reference firms: the three closest in size with and three without a word (balanced), the same six for every change (fixed), or six of any size (random size). The rule-based edit and each post-hoc editor change the plain draft after drafting (Appendix~\ref{app:prior}). The template, an unranked reference without a model, restates the input as \textit{``\{item\} increased (decreased) \{size\}\% in fiscal \{year\} compared with fiscal \{year$-$1\}.''} Six item names with a connective give it a Cause of 0.6\%.

\noindent\textbf{Other baselines.}
The drafting model runs each editor except FAVA, which uses its own model, and no editor receives the norms or flags unless stated otherwise. Table~\ref{tab:baseline_examples} shows every baseline on one test change, as a new draft or as an edit.

\begin{table*}[tp]
\centering
\caption{\textbf{Each baseline on one test change.} Mistral-Small-24B on \textit{Net sales, fiscal 1999, +107.8\%}, which the writer did not judge. Deletions are red, additions green, judgment words \jwb{orange}, and unsupported causes \icb{blue}.}
\label{tab:baseline_examples}
\small
\begin{tabular}{@{}>{\raggedright\arraybackslash}p{3.4cm}|>{\raggedright\arraybackslash}p{4.5cm}|>{\raggedright\arraybackslash}p{7.2cm}@{}}
\toprule
Method & What it changes & Output \\
\midrule
Plain prompt & Nothing & Net sales \jwb{surged} by 107.8\% from fiscal 1998 to fiscal 1999, \icb{reflecting strong market} \icb{demand and successful} \icb{strategic initiatives}. \\
\midrule
\multicolumn{3}{l}{\textit{(a) \textbf{Prompt baselines}: Prompts that state the norms or ask for faithfulness}} \\
+ Norm card & Adds the ranges of eight words & Our net sales experienced a \jwb{dramatic} increase of 107.8\% in fiscal 1999 compared with fiscal 1998. \\
+ Norm card with rate & Adds that writers rarely judge & Net sales for fiscal 1999 experienced a \jwb{dramatic} increase of 107.8\% compared with fiscal 1998. \\
+ Calibrated instruction & Judges only unusual changes & Net sales increased 107.8\% in 1999 compared to 1998. \\
+ Style instruction & Follows the 10-K style & Net sales increased 107.8\% from fiscal 1998. \\
+ Faithful instruction & Adds nothing beyond the input & Net sales increased by 107.8\% from fiscal 1998 to fiscal 1999. \\
\midrule
\multicolumn{3}{l}{\textit{(b) \textbf{Sentence baselines}: Example sentences in the prompt}} \\
+ Balanced retrieval & Three with and three without a word & Net sales \jwb{surged} by 107.8\% in fiscal 1999 compared to fiscal 1998. \\
+ Fixed sentences & The same six for every change & Net sales \jwb{surged} by 107.8\% in fiscal 1999 compared with fiscal 1998. \\
+ Random-size sentences & Six sentences of any size & Net sales for fiscal 1999 \jwb{surged} by 107.8\% compared to fiscal 1998. \\
\midrule
\multicolumn{3}{l}{\textit{(c) \textbf{Rule-based edit} after drafting}} \\
+ Rule deleting all words & Deletes or replaces every magnitude word & Net sales \delt{surged}~\inst{increased} by 107.8\% from fiscal 1998 to fiscal 1999, \icb{reflecting strong market} \icb{demand and successful} \icb{strategic initiatives}. \\
\midrule
\multicolumn{3}{l}{\textit{(d) \textbf{Post-hoc editors}}} \\
DRG & Uses words of a matched sentence & Net sales \delt{surged}~\inst{increased} by 107.8\% from fiscal 1998 to fiscal 1999, \icb{reflecting strong market} \icb{demand and successful} \icb{strategic initiatives}. \\
Self-Refine & Revises after its own feedback & Net sales \delt{surged}~\inst{increased} by 107.8\% from fiscal 1998 to fiscal 1999\deltw{, reflecting strong market demand and successful strategic initiatives}. \\
CoVe & Revises after verification questions & Net sales \delt{surged}~\inst{increased} by 107.8\% from fiscal 1998 to fiscal 1999\deltw{, reflecting strong market demand and successful strategic initiatives}. \\
FAVA & Deletes or inserts marked spans & No change \\
MiniCheck+rewrite & Rewrites unsupported drafts & Net sales \delt{surged}~\inst{increased} by 107.8\% from fiscal 1998 to fiscal 1999\deltw{, reflecting strong market demand and successful strategic initiatives}. \\
\midrule
\ourrow \ours{} & Matched sentences and rule edit & Net sales increased by 107.8\% in fiscal 1999 compared with fiscal 1998. \\
\midrule
Template (unranked) & Restates the input & Net sales increased 107.8\% in fiscal 1999 compared with fiscal 1998. \\
\bottomrule
\end{tabular}
\end{table*}

\subsection{Metrics}
\label{app:metrics}

Each metric is a share of texts (\%). For the texts $D_m$ written by one drafting LLM $m$, the metric is
\begin{equation*}
\mathrm{M}(D_m)=\frac{100}{|D_m|}\sum_{d\in D_m}\one_{\mathrm{M}}(d),
\end{equation*}
averaged over the LLMs. A text $d$ reports a change of size $s$ (\%), $W(d)$ is the set of its magnitude words in the lexicon (Section~\ref{sec:norms}), and $c(w)$ the class of $w$. Where a table is short of space, it abbreviates Placeholder as simply Placeh.
\begin{itemize}[leftmargin=*,itemsep=1pt,topsep=2pt,parsep=0pt]
\item \textbf{Judged:} $\one[W(d)\neq\emptyset]$, best near the writers.
\item \textbf{Misfit:} $\one[\exists w\in W(d)$: $s<q_{10}(w)$, $c(w)\neq$ mild, or $s>q_{90}(w)$, $c(w)\neq$ intense$]$.
\item \textbf{Cause:} $\one[d$ has a connective such as \textit{due to}$]$.
\item \textbf{Placeholder:} $\one[d$ leaves a slot for the user to fill, such as \textit{``[reason]''} or \textit{``\$X million''}$]$.
\item \textbf{Invented:} $\one[d$ states an amount that is not in the input$]$, up to its units and rounding (1.2\%).
\item \textbf{Number:} $\one[\exists x\in d\!: |x-s|\le\max(0.05,0.006s)]$, up to rounding. Writers often state a change as two amounts instead, so their Number is only 75.1\% on the same changes.
\end{itemize}
Finally, the \textbf{Gap} is $|\mathrm{Judged}-\mathrm{Judged}_{\text{writers}}|+\mathrm{Cause}+\mathrm{Placeholder}$, and \textbf{Calls} counts model calls per draft after drafting. Plain drafts give $|17.2-1.5|+77.0+8.5=101.2$. We sum the three parts with equal weight, as each is a share of drafts in points, and the main results also report them separately. The ranking is robust to adding Invented (\ours{} 1.2, template 2.1, faithful instruction 2.9), while a target of zero judgments instead of the writers' rate ranks the template (0.6) and the faithful instruction (1.5) ahead of \ours{} (1.6), at the cost of the mild words that writers use. Table~\ref{tab:metric_example} scores four drafts, and Appendix~\ref{app:metric_examples} gives more examples for each metric, one panel per metric.

\begin{table*}[t]
\centering
\caption{\textbf{Scoring four drafts.} Plain drafts for \textit{Interest expense, fiscal 2001, +9.5\%}. Writers use \textit{sharp} above $q_{10}=22.0\%$ and \textit{slight} up to $q_{90}=8.9\%$, and the Gap is $|50-1.5|+50+25=123.5$.}
\label{tab:metric_example}
\small
\setlength{\tabcolsep}{1.8pt}
\begin{tabular}{@{}l|p{7.7cm}|cccccc@{}}
\toprule
 & Draft & Judged $\approx$ & Misfit $\downarrow$ & Cause $\downarrow$ & Placeh. $\downarrow$ & Invented $\downarrow$ & Number $\uparrow$ \\
\midrule
$d_1$ & Interest expense increased \jwb{sharply} by 9.5\% in fiscal 2001, \icb{primarily} \icb{due} \icb{to} \icb{higher} \icb{borrowings} of \$12 million. & 1 & 1 & 1 & 0 & 1 & 1 \\
$d_2$ & Interest expense increased 9.5\% in fiscal 2001 \icb{due to} [reason]. & 0 & 0 & 1 & 1 & 0 & 1 \\
$d_3$ & Interest expense increased 9.5\% in fiscal 2001. & 0 & 0 & 0 & 0 & 0 & 1 \\
$d_4$ & Interest expense increased \jwb{slightly} in fiscal 2001. & 1 & 1 & 0 & 0 & 0 & 0 \\
\midrule
 & Share (\%) & 50 & 50 & 50 & 25 & 25 & 75 \\
\bottomrule
\end{tabular}
\end{table*}

\section{Examples of the Metrics}
\label{app:metric_examples}

This section shows how each metric of Appendix~\ref{app:metrics} scores single texts. All scores are computed by the same rules as in the experiments.

\noindent\textbf{Judged.}
A text counts if it has a magnitude word of any of the three classes, whether or not the word fits the change. Adverbs, adjectives on a change noun, and verbs that state the size all count (Table~\ref{tab:ex_judged}).

\begin{table*}[t]
\centering
\caption{\textbf{Examples of Judged.} A text scores 1 if it has a magnitude word of the lexicon in one of the three classes.}
\label{tab:ex_judged}
\small
\begin{tabular}{@{}p{7.4cm}|c|p{6.8cm}@{}}
\toprule
Text & Judged & Why \\
\midrule
Net income declined \textbf{slightly} to \$4.1 million. & 1 & The adverb \textit{slightly} counts as the mild word \textit{slight}. \\
Inventory grew 15.0\%, a \textbf{large increase} over the prior year. & 1 & The intense adjective \textit{large} modifies a change noun. \\
Sales \textbf{plunged} 60.0\% in fiscal 2009. & 1 & The verb \textit{plunged} states a large decrease (intense). \\
Revenue \textbf{ticked up} 1.0\%. & 1 & The verb \textit{ticked up} states a small increase (mild). \\
Costs were \textbf{somewhat} higher in fiscal 2003. & 1 & The adverb \textit{somewhat} is a mild word. \\
Operating expenses showed a \textbf{strong increase} of 30.0\%. & 0 & Evaluative \textit{strong} has no size norm, even here. \\
Revenue increased 12.0\%, driven by \textbf{robust} demand. & 0 & \textit{Robust} is evaluative and modifies demand. \\
A \textbf{significant portion} of revenue came from Europe. & 0 & \textit{Significant} modifies \textit{portion}, not a change noun. \\
There is \textbf{substantial doubt} about the going concern. & 0 & \textit{Substantial} modifies \textit{doubt}, not a change noun. \\
Revenue \textbf{increased} 40.0\% in fiscal 2010. & 0 & \textit{Increased} states only the direction of the change. \\
\bottomrule
\end{tabular}
\end{table*}

\noindent\textbf{Misfit.}
Each magnitude word is compared with the range in which the writers use it. The check is one-sided for mild and intense words, as an intense word on a very large change or a mild word on a very small change still fits. Table~\ref{tab:ex_text} gives examples of the five metrics that score a single text, one panel each, starting with Misfit in panel (a).

\begin{table*}[tp]
\centering
\caption{\textbf{Examples of the text-level metrics.} Each text scores 1 or 0. Drafts are for \textit{Interest expense, fiscal 2001, +9.5\%}, and summaries for \textit{``Interest expense increased \$1.2 million, or 9.5\%, to \$13.8 million in fiscal 2001.''}}
\label{tab:ex_text}
\small
\begin{tabular}{@{}p{7.6cm}|c|p{6.6cm}@{}}
\toprule
Text & Score & Why \\
\midrule
\multicolumn{3}{@{}p{15.3cm}@{}}{(a) \textbf{Misfit}: 1 if a moderate or intense word falls below its $q_{10}$ (inflated), or a mild or moderate word above its $q_{90}$ (deflated)} \\
\midrule
Revenue increased \textbf{significantly} by 12.0\%. & 1 & \textit{significant}, 17.5--256.3: inflated, as $12.0<17.5$. \\
Revenue increased \textbf{significantly} by 300.0\%. & 0 & \textit{significant}, 17.5--256.3: intense, so no upper bound. \\
Costs rose \textbf{slightly} by 12.0\%. & 1 & \textit{slight}, 0.3--8.9: deflated, as $12.0>8.9$. \\
\midrule
\multicolumn{3}{@{}p{15.3cm}@{}}{(b) \textbf{Cause}: 1 if a causal connective starts a clause} \\
\midrule
Interest expense increased 9.5\% \textbf{due to} higher borrowings. & 1 & \textit{Due to} starts a causal clause. \\
Interest expense increased 9.5\% compared with fiscal 2000. & 0 & A comparison of periods is not a cause. \\
Interest expense rose 9.5\% after the company issued notes. & 0 & A time clause with no connective, missed by the rule. \\
\midrule
\multicolumn{3}{@{}p{15.3cm}@{}}{(c) \textbf{Placeholder}: 1 if the text leaves a slot to fill} \\
\midrule
Interest expense increased 9.5\% due to \textbf{[reason]}. & 1 & A slot in square brackets. \\
Interest expense increased by \textbf{\$X million}, or 9.5\%. & 1 & An amount left as \textit{\$X}. \\
Interest expense increased 9.5\% to \$12 million. & 0 & A stated amount, counted by Invented instead. \\
\midrule
\multicolumn{3}{@{}p{15.3cm}@{}}{(d) \textbf{Invented}: 1 if the text states an amount not in its input} \\
\midrule
\textit{Drafting:} Interest expense increased 9.5\% to \textbf{\$12 million}. & 1 & Any amount is invented, as the input has none. \\
\textit{Summary:} Interest expense rose 9.5\% to \textbf{\$13,800 thousand}. & 0 & The same amount in other units. \\
\textit{Summary:} Interest expense rose 9.5\% to \textbf{\$14 million}. & 1 & Rounding to \$14 million exceeds the 1.2\% tolerance. \\
\midrule
\multicolumn{3}{@{}p{15.3cm}@{}}{(e) \textbf{Number}: 1 if the text states the size of the change ($s=9.5$)} \\
\midrule
Interest expense increased \textbf{9.5\%}. & 1 & The exact size. \\
Interest expense increased \textbf{approximately 10\%}. & 1 & 10 is 9.5 rounded to a whole number. \\
Interest expense increased \textbf{9\%}. & 0 & 9 is neither within the tolerance nor 9.5 rounded. \\
\bottomrule
\end{tabular}
\end{table*}

\noindent\textbf{Cause.}
A text counts if it has a causal clause, found by a connective like \textit{due to}. Panel (b) shows clauses that the rule counts and those that it does not.

\noindent\textbf{Placeholder.}
A text counts if it leaves a slot for the user to fill, such as a bracket like \textit{[reason]} or an amount written as \textit{\$X million} (panel c).

\noindent\textbf{Invented.}
A text counts if it states an amount not in its input. In drafting, any amount is invented, while a summary may restate the source numbers in other units or their differences (panel d).

\noindent\textbf{Number.}
A text counts if it states the size of the change in its input, allowing other decimals and rounding to a whole number. Panel (e) shows forms of the size that count and forms that do not.

\section{Extraction of (Word, Change) Pairs}
\label{app:extraction}

\begin{table}[t]
\centering
\caption{\textbf{Statistics of the 10-K norm corpus.} Changes in MD\&A sections whose size can be extracted.}
\label{tab:corpus}
\small
\begin{tabular}{@{}l|r@{}}
\toprule
Statistic & Value \\
\midrule
Filings & 220{,}375 \\
Firms & 38{,}009 \\
\quad with an extracted change & 20{,}916 \\
Years & 1993--2020 \\
Changes & 2{,}459{,}128 \\
\quad with a magnitude word & 18{,}484 \\
Lemmas found in extracted pairs & 34 \\
\bottomrule
\end{tabular}

\vspace{8pt}
\caption{\textbf{Held-out audit of the extracted pairs.} Precision with a 95\% Wilson interval.}
\label{tab:audit}
\small
\begin{tabular}{@{}l|cc@{}}
\toprule
Corpus & Correct & 95\% CI \\
\midrule
10-K MD\&A & 94/100 & 0.875--0.972 \\
Financial news & 50/50 & 0.929--1.000 \\
General news & 49/50 & 0.895--0.996 \\
\midrule
All & 193/200 & 0.930--0.983 \\
\bottomrule
\end{tabular}
\end{table}

We extract a (word, change) pair from each reported change in the steps below. Table~\ref{tab:corpus} summarizes the resulting corpus of changes, and a held-out audit checks the pairs.

\noindent\textbf{Sentences and change words.}
We split each MD\&A section into sentences and keep those that contain a change verb (e.g., \textit{increased}, \textit{grew}, \textit{rose}, \textit{gained}, \textit{decreased}, \textit{declined}, \textit{fell}, \textit{dropped}, \textit{reduced}) or a change noun (e.g., \textit{increase}, \textit{growth}, \textit{rise}, \textit{gain}, \textit{decrease}, \textit{decline}, \textit{drop}, \textit{reduction}).

A magnitude word is linked to the change only when it is adjacent to it (e.g., \textit{increased slightly}, \textit{slightly increased}, \textit{a slight increase}). Verbs that themselves express the size of a change, such as \textit{surged} or \textit{edged up}, are their own magnitude words. We drop a candidate when the next word shows that the word does not judge the change itself (e.g., \textit{significantly offset}, \textit{slightly higher than}).

\noindent\textbf{Lexicon.}
The lexicon of \ours{} groups the words by lemma, where an adverb and its adjective share one entry (e.g., \textit{slightly} and \textit{slight}) and verbs are listed by their base form. \textit{Mild} words are \textit{slight}, \textit{marginal}, \textit{modest}, \textit{minimal}, \textit{small}, \textit{minor}, \textit{somewhat}, and the verbs \textit{edge}, \textit{tick}, \textit{inch}, \textit{creep}, \textit{dip}, \textit{slip}, \textit{ease}, and \textit{nudge}. \textit{Moderate} words are \textit{moderate}, \textit{noticeable}, and the verbs \textit{climb} and \textit{slide}.

\textit{Intense} words are \textit{significant}, \textit{substantial}, \textit{considerable}, \textit{sharp}, \textit{dramatic}, \textit{steep}, \textit{large}, \textit{material}, \textit{meaningful}, \textit{marked}, \textit{massive}, \textit{tremendous}, \textit{remarkable}, \textit{sizable}, \textit{huge}, and the verbs \textit{surge}, \textit{soar}, \textit{jump}, \textit{rally}, \textit{spike}, \textit{skyrocket}, \textit{rocket}, \textit{leap}, \textit{plunge}, \textit{plummet}, \textit{tumble}, \textit{slump}, \textit{sink}, \textit{dive}, \textit{crash}, \textit{tank}, \textit{balloon}, and \textit{shrink}. The evaluative words \textit{strong}, \textit{robust}, and \textit{notable} are found but neither counted in the judgment rate nor flagged as misfits.

\noindent\textbf{Size of the change.}
We read the size from the words that follow the change, in a fixed order. 1)~A change in a ratio, such as \textit{to 53\% from 52\%}, gives a change in percentage points. 2)~A percentage, such as \textit{by 3\%} or \textit{(3\%)}, gives a relative change, unless the percentage is a level (\textit{30\% of revenues}) or belongs to another quantity. 3)~Two amounts, such as \textit{from \$A to \$B} or \textit{to \$B compared to \$A}, give a relative change. 4)~A difference and a new level, such as \textit{by \$D to \$B}, do too. For amounts, we also keep the interval implied by the rounding of the stated numbers, and drop changes whose interval exceeds half of the change. The main norms use only the relative changes of these four forms.

\begin{table}[t]
\centering
\caption{\textbf{Usage norms of the reference firms.} Change size (\%) at the 10th, 50th, and 90th percentiles and the mean, for words with at least 30 uses and for each class, which rarer words use. Shading marks the class.}
\label{tab:norms_words}
\small
\setlength{\tabcolsep}{6.4pt}
\begin{tabular}{@{}l|ccc|c|c@{\hspace{\tabcolsep}}}
\toprule
Word & $q_{10}$ & $q_{50}$ & $q_{90}$ & Mean & Uses \\
\midrule
\multicolumn{6}{l}{\textit{Mild}} \\
\rowcolor{blue!4} \textit{marginal(ly)} & 0.3 & 1.6 & 6.2 & 3.3 & 119 \\
\rowcolor{blue!4} \textit{modest(ly)} & 0.4 & 2.2 & 8.1 & 4.6 & 209 \\
\rowcolor{blue!4} \textit{slight(ly)} & 0.3 & 1.8 & 8.9 & 4.7 & 2{,}573 \\
\rowcolor{blue!4} \textit{small} & 0.3 & 2.4 & 9.0 & 8.8 & 41 \\
\midrule
\rowcolor{blue!4} Mild class & 0.3 & 1.9 & 8.8 & 4.7 & 3{,}009 \\
\midrule
\multicolumn{6}{l}{\textit{Moderate}} \\
\rowcolor{blue!10} \textit{moderate(ly)} & 0.8 & 3.1 & 9.9 & 6.0 & 40 \\
\rowcolor{blue!10} \textit{climbed} & 3.0 & 21.8 & 66.4 & 35.2 & 162 \\
\midrule
\rowcolor{blue!10} Moderate class & 1.9 & 15.0 & 56.0 & 28.7 & 211 \\
\midrule
\multicolumn{6}{l}{\textit{Intense}} \\
\rowcolor{blue!18} \textit{jumped} & 4.8 & 48.0 & 143.1 & 77.5 & 45 \\
\rowcolor{blue!18} \textit{significant(ly)} & 17.5 & 61.8 & 256.3 & 108.3 & 1{,}102 \\
\rowcolor{blue!18} \textit{dramatic(ally)} & 13.2 & 70.0 & 205.3 & 111.3 & 58 \\
\rowcolor{blue!18} \textit{sharp(ly)} & 22.0 & 55.9 & 361.8 & 125.0 & 59 \\
\rowcolor{blue!18} \textit{substantial(ly)} & 26.3 & 75.3 & 341.1 & 143.5 & 255 \\
\rowcolor{blue!18} \textit{large} & 15.6 & 73.7 & 362.0 & 161.8 & 33 \\
\midrule
\rowcolor{blue!18} Intense class & 16.0 & 61.8 & 282.4 & 111.0 & 1{,}671 \\
\bottomrule
\end{tabular}
\end{table}

\noindent\textbf{Audit.}
We improved the rules on a development sample and then audited 200 pairs drawn after the rules were fixed, with 100 from 10-K filings and 50 from each of two news corpora, financial\footnote{\url{https://huggingface.co/datasets/ashraq/financial-news-articles}} and general\footnote{\url{https://huggingface.co/datasets/vblagoje/cc_news}}. A pair is correct if the size matches the change judged by the word in the sentence. In Table~\ref{tab:audit}, most remaining errors come from a number of another quantity in the same sentence.

\section{Details of \ours{}}
\label{app:steps}

\subsection{Building the Usage Norms}
\label{app:norms}

\noindent\textbf{Pairs.}
From each sentence reporting a change, Appendix~\ref{app:extraction} gives the magnitude word, if any, and the size of the change, kept up to 1{,}000\%.

\noindent\textbf{Ranges.}
For each word with at least 30 distinct sentences in the reference firms, the usage norm spans the 10th to the 90th percentile of its change sizes. A rarer word takes the range of its class, mild, moderate, or intense, pooling all its words. Table~\ref{tab:norms_words} lists the norms. As the ranges of mild and intense words barely overlap, a word on a change of the other end is easy to flag, while a word near the edge of its range is marked for review only, as the Misfit examples in Appendix~\ref{app:metric_examples} show.

\begin{table}[t]
\centering
\caption{\textbf{Size-matched sentences for one change.} Six sentences that Step~1 adds to the prompt for an increase of 3\%, all from reference firms, with each size \szb{shaded}.}
\label{tab:retrieval_example}
\small
\setlength{\tabcolsep}{4pt}
\begin{tabular}{@{}p{0.93\columnwidth}@{}}
\toprule
\textbf{Input:} Life insurance premium, fiscal 2007 compared with fiscal 2006, \szb{+3\%} \\
\specialrule{0.4pt}{1pt}{0pt}\specialrule{0.4pt}{1.5pt}{1pt}
On a constant currency basis, revenue increased \szb{3 percent} over fiscal year 2008. \\
\specialrule{0.2pt}{1pt}{1pt}
At constant exchange rates, international segment sales increased \szb{3 percent} for the year. \\
\specialrule{0.2pt}{1pt}{1pt}
Our number of general and administrative employees increased \szb{3\%} during 2013. \\
\specialrule{0.2pt}{1pt}{1pt}
Sales to OEMs increased by \szb{3\%}, while replacement sales were flat on a year-over-year basis. \\
\specialrule{0.2pt}{1pt}{1pt}
Gross ton-miles and revenue ton-miles increased \szb{3\%}. \\
\specialrule{0.2pt}{1pt}{1pt}
Total revenue would have increased by \szb{3\%} in fiscal 2002 if exchange rates had been constant. \\
\bottomrule
\end{tabular}
\end{table}

\noindent\textbf{Writers' rate.}
The reference for the judgment rate is the share of writers' sentences with any magnitude word, counted with the same lexicon as the drafts on 28K sentences of the test firms, for each of seven size bins. Writers judge most often on the smallest changes, mostly with a mild word, and report all other sizes with the number alone in about 99 of 100 sentences. The extracted pairs give a lower share, as they count only attached words.

\subsection{Step 1: Size-Matched Retrieval}
\label{app:retrieval}

\noindent\textbf{Pool.}
The pool holds 400K sentences of the reference firms that report a change of a plain line item in fewer than 300 characters. No test firm enters the pool, and near-duplicates across firms do not change the main results (Appendix~\ref{app:results}).

\noindent\textbf{Clean sentences.}
We keep sentences that state the change as a percentage, contain no dollar sign or other number except a year, and no causal cue such as \textit{due to}. A drafting model copies what it sees, so amounts or causes in them lead it to invent amounts, leave placeholders, or add its own causes.

\noindent\textbf{Selection and prompt.}
For each change, \ours{} takes the six clean sentences of the same direction whose size is closest on a log scale, so 3\% is close to 3.5\% but far from 30\%. It does not filter by magnitude words, so the sentences keep the natural mix of the writers, where about one in a hundred carries a word (Table~\ref{tab:retrieval_example}). The sentences follow the instruction as examples of how writers report a change of similar size (Appendix~\ref{app:prompts}).

\noindent\textbf{Lightweight variant.}
\ours{} (style) replaces the six sentences with one sentence in the instruction: \textit{``Follow the style of 10-K filings, which usually report a change with its number alone and rarely add words such as slightly or significantly.''} It adds only 29 tokens to the drafting prompt.

\subsection{Step 2: Checking Words and Causes}
\label{app:check}

\noindent\textbf{Finding judgment words.}
\ours{} finds magnitude words with the lexicon of Appendix~\ref{app:extraction}, 78 forms of 55 lemmas, three of them evaluative, 34 of which occur in the extracted pairs. An adverb or a verb such as \textit{surged} counts wherever it appears, while an adjective counts only when a change noun follows within two words, so \textit{a significant portion of revenue} is skipped. A word is skipped when \textit{no}, \textit{not}, \textit{without}, \textit{nor}, or \textit{never} occurs up to three words before it in the same clause, and evaluative words such as \textit{strong} are never flagged or counted.

\noindent\textbf{Misfit flag.}
Each word is compared with the size of the change in the input, not with a number in the draft. A word with fewer than 30 uses takes the range of its class, as \textit{surged} does in Appendix~\ref{app:norms}, and the lower bound of an intense word is never below that of the intense class as a whole.

\noindent\textbf{Finding causes.}
A cause clause starts at a causal connective, namely \textit{due to}, \textit{driven by}, \textit{attributable to}, \textit{attributed to}, \textit{as a result of}, \textit{resulting from}, \textit{owing to}, \textit{because of}, \textit{caused by}, \textit{led by}, \textit{fueled by}, \textit{thanks to}, or \textit{reflecting}, or at an interpretive verb such as \textit{indicating} or \textit{highlighting}. The connective may also follow an adverb such as \textit{primarily} or \textit{mainly}, or a relative clause such as \textit{which was}.

\noindent\textbf{Support from the input.}
When the input gives causes, as notes do, \ours{} keeps the content words of the clause, stems them, and computes the share of stems that occur in the input. The clause is supported when this share reaches 0.58, the threshold with the highest balanced accuracy (0.90) on 621 drafts of the reference firms.

\noindent\textbf{Limits.}
The connective list misses causes that start with other verbs, such as \textit{demonstrating}, or that sit in the subject (Appendix~\ref{app:examples}). A clause that mixes a supported and an unsupported cause is kept or cut as a whole. Such clauses are rare, and splitting them into single causes does not improve the edit. Appendix~\ref{app:cause} validates the cause check.

\subsection{Step 3: Editing Flagged Texts}
\label{app:edit}

\begin{table*}[t]
\centering
\caption{\textbf{Rule edit of Step 3 on diverse cases.} Each draft shows the output of the code, with deleted text in red strikethrough and added text in green, and with judgment words and cause clauses shaded as in Figure~\ref{fig:teaser}. Review: the cut is skipped and the clause is flagged for the preparer.}
\label{tab:edit_examples}
\small
\setlength{\tabcolsep}{4pt}
\begin{tabular}{@{}l|l|c|l@{}}
\toprule
Case & Draft and its edit & Size & Flag \\
\midrule
\multicolumn{4}{l}{\textit{Judgment words}} \\
\multirow{4}{*}{Adverb} & Revenue increased \delt{\jwb{significantly}} by 3\%. & 3\% & Inflated \\
 & Revenue increased \jwb{significantly} by 60\%. & 60\% & None \\
 & Revenue increased \delt{\jwb{slightly}} by 40\%. & 40\% & Deflated \\
 & Revenue increased \delt{\jwb{moderately}} by 25\%. & 25\% & Deflated \\
\cmidrule{1-4}
\multirow{2}{*}{Adjective} & Sales saw \delt{\jwb{a significant}} \inst{an} increase of 3\%. & 3\% & Inflated \\
 & Sales saw \delt{\jwb{a slight}} \inst{an} increase of 45\%. & 45\% & Deflated \\
\cmidrule{1-4}
\multirow{4}{*}{Verb} & Revenue \delt{\jwb{surged}} \inst{increased} 3\%. & 3\% & Inflated \\
 & Revenue \jwb{surged} 80\%. & 80\% & None \\
 & Costs \delt{\jwb{plunged}} \inst{decreased} 4\%. & 4\% & Inflated \\
 & Costs \delt{\jwb{edged up}} \inst{increased} 40\%. & 40\% & Deflated \\
\midrule
\multicolumn{4}{l}{\textit{Words that are not flagged}} \\
Evaluative word & Revenue showed strong growth of 3\%. & 3\% & None \\
No change noun & A significant portion of revenue increased 3\%. & 3\% & None \\
Negation & Revenue did not increase significantly, rising 3\%. & 3\% & None \\
\midrule
\multicolumn{4}{l}{\textit{Cause clauses}} \\
Unsupported cause & Revenue increased 3\% in fiscal 2016\delt{\icb{, driven by strong demand}}. & 3\% & Cause \\
Placeholder & Revenue increased 3\% in fiscal 2016\delt{\icb{, primarily due to [reason]}}. & 3\% & Cause \\
Word in a cause & Revenue increased \delt{\jwb{significantly}} by 3\% in 2016\delt{\icb{, reflecting a significant rise}}. & 3\% & Both \\
\midrule
\multicolumn{4}{l}{\textit{Safety rule}} \\
Short but complete & Revenue increased 3\%\delt{\icb{, driven by demand}}. & 3\% & Cause \\
Cause first & \icb{Due to strong demand}, revenue rose 3\% in 2016. & 3\% & Review \\
Cause as predicate & Revenue growth was \icb{due to a modest increase in rent}. & -- & Review \\
\bottomrule
\end{tabular}
\end{table*}

\noindent\textbf{Rules.}
The edit first cuts unsupported cause clauses, as many judgment words and placeholders sit inside them, and then edits the misfit words that remain. A clause is cut from its connective, with any comma or \textit{and} before it, to the end of the sentence. A flagged adverb is deleted, a flagged adjective is deleted with the article fixed, so \textit{a significant increase} becomes \textit{an increase}, and a flagged verb becomes \textit{increased} or \textit{decreased}, with a particle such as \textit{up} in \textit{edged up}. Table~\ref{tab:edit_examples} shows each rule on diverse cases, as the code outputs them.

\noindent\textbf{Safety rule.}
A cut is skipped when it would remove the number of the change, or when the remainder is shorter than five words and has no change verb, as in \textit{Revenue growth was}. A short remainder that still reports the change, such as \textit{Revenue increased 3\%}, is cut, as an invented cause misleads readers more than a short sentence. A skipped clause keeps its flag for review, and LLM judges find almost all edited drafts complete sentences (Appendix~\ref{app:results}).

\begin{table}[t]
\centering
\caption{\textbf{Writers on good and bad news.} Logistic regression of an intense over a mild word on 7{,}502 changes. An intense word becomes likely above 17.4\% for good and 26.0\% for bad news.}
\label{tab:asym}
\small
\setlength{\tabcolsep}{8.0pt}
\begin{tabular}{@{}l|ccc@{}}
\toprule
 & Coefficient & SE & $p$ \\
\midrule
Log size of change & 2.585 & 0.101 & $<$0.001 \\
Bad news & $-$1.040 & 0.164 & $<$0.001 \\
\bottomrule
\end{tabular}
\end{table}

\noindent\textbf{No model call.}
All rules are string operations that take well under a millisecond per draft on a CPU. As the edit never writes new text, it cannot add a judgment, cause, or amount, which a model asked to rewrite the draft may do (Appendix~\ref{app:prior}).

\noindent\textbf{Alternatives.}
A \textit{revision} returns each flagged draft to the model once with the word's typical range, at the cost of one call (Appendix~\ref{app:results}). A \textit{slot} leaves \textit{``[explain driver]''} in place of a cut cause, for the preparer (Appendix~\ref{app:deploy}). With retrieval, 20.5\% of the drafts get one, as 0.4\% keep their clause (Table~\ref{tab:components}).

\section{Further Analysis of Writers}
\label{app:writers}

\noindent\textbf{Good- and bad-news labels.}
To decide whether a change is good or bad news for the firm in Table~\ref{tab:asym}, we combine the direction of the change with the kind of item. A rise in revenue, income, or margin is good, while a rise in cost, expense, loss, or debt is bad, and the opposite holds for a fall.

Market prices that the firm does not control, such as commodity prices or exchange rates, are left unlabeled. A held-out audit of 60 labels finds 57 correct (95\% Wilson interval 0.863--0.983).

\begin{table*}[t]
\centering
\caption{\textbf{Three wordings of the instruction.} Texts (\%) with a judgment word (Judged, best near the writers) or an unsupported cause, and the Gap, over the 17 LLMs, for the main instruction (P0) and its paraphrases P1 and P2.}
\label{tab:wordings}
\small
\setlength{\tabcolsep}{4pt}
\begin{tabular}{@{}l|ccc|ccc|ccc@{}}
\toprule
 & \multicolumn{3}{c|}{P0} & \multicolumn{3}{c|}{P1} & \multicolumn{3}{c}{P2} \\
\cmidrule(lr){2-4}\cmidrule(lr){5-7}\cmidrule(lr){8-10}
 & Judged $\approx$ & Cause $\downarrow$ & Gap $\downarrow$ & Judged $\approx$ & Cause $\downarrow$ & Gap $\downarrow$ & Judged $\approx$ & Cause $\downarrow$ & Gap $\downarrow$ \\
\midrule
Plain prompt & 17.2 & 77.0 & 101.2 & 4.2 & 23.7 & 34.7 & 8.9 & 33.1 & 50.6 \\
+ Faithful instruction & \second{0.3} & \second{0.6} & \second{2.3} & 0.0 & 0.6 & \second{4.9} & 0.1 & 0.5 & \second{5.4} \\
\midrule
Retrieval & 4.0 & 20.9 & 24.8 & \best{0.6} & 5.0 & 6.3 & \best{0.9} & 4.9 & 5.8 \\
\ourrow \ours{} & \best{1.0} & \best{0.4} & \best{1.1} & \second{0.2} & \second{0.4} & \best{1.9} & \second{0.2} & \second{0.3} & \best{1.8} \\
\ourrow \ours{} (style) & 0.0 & \best{0.4} & 4.5 & 0.0 & \best{0.3} & 7.2 & 0.0 & \best{0.2} & 6.6 \\
\specialrule{0.4pt}{1pt}{1pt}\specialrule{0.4pt}{0pt}{1pt}
\writerrow 10-K writers & 1.5 & -- & -- & 1.5 & -- & -- & 1.5 & -- & -- \\
Template (unranked) & 0.0 & 0.6 & 2.1 & 0.0 & 0.6 & 2.1 & 0.0 & 0.6 & 2.1 \\
\bottomrule
\end{tabular}
\end{table*}

\noindent\textbf{Controls.}
To test whether the bad-news effect stems from other factors, we add controls to the bad-news regression: the direction of change, coarse rounding of stated numbers (e.g., \textit{about \$10 million}), the period before and after 2002, when the Sarbanes-Oxley Act tightened reporting rules, and fixed effects for the kind of item (e.g., revenue or expense). In every specification, the coefficient of bad news stays between $-1.13$ and $-1.04$, and it is significant at $p<0.001$ in all four. The size of the change alone separates most mild from intense words (pseudo-$R^2$ of 0.77 to 0.80), and coarse rounding of the stated numbers clearly lowers the chance of using an intense word.

\section{Prompts}
\label{app:prompts}

All models receive the same system message, \textit{``You are an experienced financial reporting specialist at a US public company.''}, followed by the user message below. The fields in braces are filled from the line item. Models without a system role receive the system message at the start of the user message.

\noindent\textbf{Drafting input.}
Every drafting prompt, from P0 to P2, starts with the same three lines below:
\begin{quote}\small\ttfamily
Line item: \{subject\}\\
Period: fiscal \{year\} compared with fiscal \{year$-$1\}\\
Change: \{sign\}\{size\}\%
\end{quote}

\noindent\textbf{Drafting instructions.}
The main instruction (P0) and its two paraphrases (P1, P2) are:
\begin{itemize}[leftmargin=0.3cm,itemsep=1pt]
\item P0: \textit{Write one sentence describing this change for the Management's Discussion and Analysis (MD\&A) section of the company's annual report on Form 10-K. Reply with the sentence only.}
\item P1: \textit{Draft the sentence that reports this change in the company's 10-K MD\&A. Output only that sentence.}
\item P2: \textit{In a single sentence, describe this year-over-year change as it would appear in the company's annual report (Form 10-K). Return just the sentence.}
\end{itemize}

Table~\ref{tab:wordings} reports each wording. Plain drafts judge 4.2\% to 17.2\% of the changes and add a cause in 23.7\% to 77.0\%, depending on the wording, while \ours{} judges at most 1.0\% and adds a cause in at most 0.4\% under every wording. Averaged over the three wordings, the Gap is 1.6 for \ours{}, 2.1 for the template, and 4.2 for the faithful instruction, whose placeholders rise under P1 and P2. Greedy decoding of the main wording gives nearly the same plain drafts as sampling at temperature 0.7 (16.5\% of the drafts judged and 77.0\% with a cause, against 17.2\% and 77.0\%).

\noindent\textbf{Norm card and retrieval.}
Both are placed before the drafting input. The norm card lists, for eight common words, the $q_{10}$ to $q_{90}$ range and the median of the change in the reference firms, in the form \textit{``- significant: 17.51\% to 256.27\% (median 61.84\%)''}, and the retrieved sentences follow the line \textit{``Sentences from other companies' 10-K filings that describe changes of a similar size:''}.

\noindent\textbf{Added instructions.}
The norm card with rate adds \textit{``Most changes in 10-K filings (about 99 in 100) are reported without any magnitude word.''} Three instructions are appended to the drafting instruction:
\begin{itemize}[leftmargin=*,itemsep=0pt,topsep=2pt,parsep=0pt]
\item Calibrated: \textit{``Follow the style of 10-K filings: most sentences report a change with its number alone. Add a word such as slightly or significantly only when the change is unusually small or large, and choose a word that fits its size.''}
\item Style: \textit{``Follow the style of 10-K filings, which usually report a change with its number alone and rarely add words such as slightly or significantly.''}
\item Faithful: \textit{``Report only what the input gives: do not add a cause for the change or any word that judges how large it is.''}
\end{itemize}

\noindent\textbf{Summarization.}
The summarization prompt is \textit{``Sentence from a company's 10-K MD\&A section: ``\{sentence\}'' Summarize this sentence for a short investor update in at most 15 words. Reply with the summary only.''} With size-matched retrieval, the sentences are placed before it as for drafting.

\noindent\textbf{\ours{} revision.}
A flagged text is returned to the model in the same conversation with the note: \textit{``In 10-K filings, ``\{word\}'' is typically used for changes between \{$q_{10}$\}\% and \{$q_{90}$\}\% (median \{median\}\%). This change is \{size\}\%. For a change of this size, 10-K writers typically use \{words\}, or no magnitude word. Replace the word ``\{word\}'' with one of these options or remove it. Do not change anything else in the sentence. Reply with the revised sentence only.''}

If no common word fits the size, the note says \textit{``Remove the word ``\{word\}'' and do not add another magnitude word.''} For summaries, \textit{sentence} becomes \textit{summary}. Revisions use greedy decoding.

\noindent\textbf{LLM judge.}
For detection, the judge is prompted with \textit{``Sentence from a 10-K MD\&A section: ``\{sentence\}'' In this sentence, does the word ``\{word\}'' fit the actual size of the change, as an experienced 10-K writer would use it? Answer yes or no.''} For LLM-written drafts, the judge receives the draft and the actual change instead, with the question \textit{``Does the wording describe the size of the change accurately, as an experienced 10-K writer would? Answer yes or no.''} We read the probabilities of \textit{yes} and \textit{no} from the first output token and use the normalized $P(\textit{no})$ as the misfit score of each draft.

\section{Validating the Cause Check}
\label{app:cause}

\begin{table}[t]
\centering
\caption{\textbf{Finding causes.} Precision (Prec.), recall, and F1 against the judges' gold on 740 test drafts, for any cause (top, with span F1) and for unsupported causes (bottom, with the false positive rate).}
\label{tab:cause_detect}
\small
\setlength{\tabcolsep}{1.0pt}
\begin{tabular}{@{}l|cccc@{}}
\toprule
Any cause & Prec. $\uparrow$ & Recall $\uparrow$ & F1 $\uparrow$ & Span $\uparrow$ \\
\midrule
Qwen2.5-7B span & \second{0.675} & \second{0.972} & \second{0.797} & \second{0.828} \\
Llama-3.1-8B span & 0.538 & \best{0.994} & 0.698 & 0.736 \\
\ourrow \ours{} rule & \best{0.896} & 0.946 & \best{0.920} & \best{0.899} \\
\midrule
Unsupported cause & Prec. $\uparrow$ & Recall $\uparrow$ & F1 $\uparrow$ & FPR $\downarrow$ \\
\midrule
Qwen2.5-7B span & 0.500 & 0.680 & 0.576 & 0.247 \\
Connective rule & 0.564 & \best{0.959} & 0.711 & 0.269 \\
\ \ + MiniCheck & \second{0.813} & \second{0.863} & \second{0.837} & \second{0.072} \\
\ \ + Bespoke-MiniCheck-7B & \best{0.880} & 0.822 & \best{0.850} & \best{0.041} \\
\ourrow \ \ + Lexical (\ours{}) & 0.802 & \second{0.863} & 0.831 & 0.077 \\
\bottomrule
\end{tabular}
\end{table}

\begin{table}[t]
\centering
\caption{\textbf{Editing drafts with notes.} Left: Drafts (\%) keeping an unsupported (Unsup.) or supported (Sup.) cause; a full cut skips causes it cannot remove cleanly. Right: Judged and Number of each edit with notes.}
\label{tab:cause_variant}
\small
\setlength{\tabcolsep}{1.0pt}
\begin{tabular}{@{}l|cc|cc@{}}
\toprule
 & Unsup. $\downarrow$ & Sup. $\uparrow$ & Judged $\approx$ & Number $\uparrow$ \\
\specialrule{0.4pt}{1pt}{1pt}\specialrule{0.4pt}{0pt}{1pt}
\writerrow 10-K writers & -- & -- & 1.5 & -- \\
\midrule
Word edit only & 10.6 & \best{100} & 4.1 & 99.8 \\
Cut all causes & \best{0.6} & 4.8 & \second{1.2} & 99.8 \\
MiniCheck support & \second{2.4} & 89 & 2.0 & 99.8 \\
\ourrow \ours{} (lexical) & 2.7 & \second{94} & \best{1.6} & 99.8 \\
\bottomrule
\end{tabular}
\end{table}

\noindent\textbf{Gold causes.}
We sample 740 test drafts evenly over the 10-model subset and the settings with and without notes, summaries, and paragraphs. Three open LLM judges, Qwen2.5-32B, Mistral-Small-24B, and gpt-oss-20B, mark each cause and label it as supported or unsupported by the input. They agree well (Fleiss $\kappa$ of 0.88 for the presence and 0.81 for the label), and the majority is gold.

\noindent\textbf{Finding causes.}
Table~\ref{tab:cause_detect} compares the connective rule with 7B and 8B LLMs that mark the cause span. The rule is more precise, and its misses mostly use connectives outside its list, such as \textit{consistent with}. Deleting the spans marked by an LLM leaves a broken sentence in about half of the edited drafts, whereas the cuts of the rule almost never do.

\noindent\textbf{Supported causes.}
The rule alone cannot tell whether the input supports a cause, which matters only when the input gives causes. Adding the lexical support test of Section~\ref{sec:causes} detects unsupported causes about as well as MiniCheck and Bespoke-MiniCheck-7B, without a model call.

\noindent\textbf{Editing with notes.}
Table~\ref{tab:cause_variant} shows the edit on drafts written from notes. Cutting every cause removes supported causes with the unsupported ones, while the support test keeps most supported causes. Over the 17 LLMs, it brings \ours{} with notes close to the writers' rate (1.6\% vs.\ 1.5\%) and keeps the number as often as the plain drafts.

\noindent\textbf{Regulatory words.}
In real filings, \textit{material} and \textit{materially} mostly carry their legal sense, as in \textit{could materially affect}. On 500 occurrences labeled by the same judges, a short rule that skips these patterns raises the precision of the word check on held-out sentences without lowering its recall. Such uses are almost absent from LLM drafts and from the writer sentences of the 980 test items.

\section{Filings after LLMs}
\label{app:wild}

\begin{table*}[t]
\centering
\caption{\textbf{10-K practice by period.} Share of change sentences (\%) in real filings, adjusted to a fixed mix of change sizes, the shift after ChatGPT within the same firms (points, 95\% interval), and plain LLM drafts.}
\label{tab:wild_year}
\small
\setlength{\tabcolsep}{8pt}
\begin{tabular}{@{}l|ccc|c|c@{\hspace{\tabcolsep}}}
\toprule
 & 2016--20 & 2021--22 & 2023--25 & Shift & LLM \\
\midrule
Magnitude word & 1.12 & 1.14 & 1.03 & $-$0.14 [$-$0.24, $-$0.05] & 17.2 \\
Intense word & 0.53 & 0.65 & 0.53 & $-$0.10 [$-$0.18, $-$0.03] & 12.1 \\
Cause clause & 29.8 & 29.3 & 30.5 & +1.27 [+0.75, +1.79] & 77.0 \\
\bottomrule
\end{tabular}
\end{table*}

If writers had started drafting filings with LLMs, the filings would drift toward the drafts and stop being a useful reference. We therefore repeat the writer measurement on 70.5K 10-K filings from 2016 to 2025, about 21K after the release of ChatGPT, using EDGAR-CORPUS and its public extension.\footnote{\url{https://huggingface.co/datasets/seanm924/edgar-corpus-extended}} We hold the change size fixed by weighting the size bins as in the drafting items.

As shown in Table~\ref{tab:wild_year}, magnitude words in filings do not rise after 2022 and even fall slightly within the same firms, including intense words. Causal clauses grow by about one point, following a trend that starts before 2022, and both stay far below the drafts. The two data sources agree around 2020 in coverage, length, and rates.

\section{Intervals and Trivial Rules}
\label{app:stats}

\begin{figure}[t]
\centering
\includegraphics[width=\columnwidth]{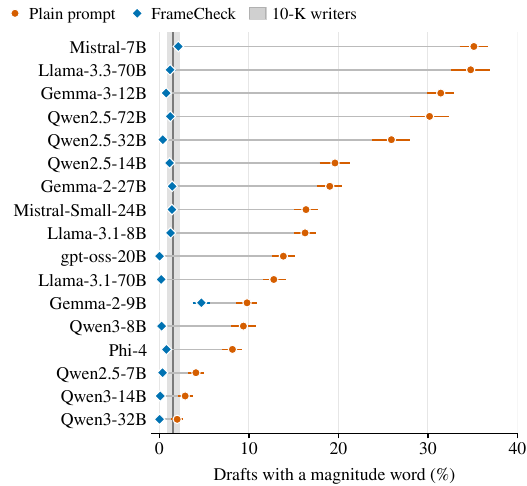}
\caption{\textbf{Judgment rate for all 17 drafting LLMs.} Plain drafts and \ours{} with 95\% bootstrap intervals, against the writers on the same changes (gray).}
\label{fig:models_all}
\end{figure}

\noindent\textbf{Bootstrap.}
We resample the 980 changes within each size bin and direction, keeping all drafts of a change and the writer's sentence together, and report 95\% intervals. The 95\% interval of \ours{} minus the writers on the same changes contains zero, while on the 28K writer sentences \ours{} judges slightly less often (1.0\% vs.\ 1.5\%). Single-model intervals are narrower, and most exclude the writers' rate (Figure~\ref{fig:models_all}).

\noindent\textbf{Intervals for the main results.}
Table~\ref{tab:ci_main} gives the 95\% intervals for the main results of Section~\ref{sec:res_audit}. The Gap of \ours{} (0.7 to 1.4) stays below the template (1.7 to 2.6), the faithful instruction (2.0 to 2.8), and every other baseline. When the writers' rate on the same changes is resampled as well, the paired difference to \ours{} remains positive for the template (1.0, from 0.5 to 1.5) and the faithful instruction (1.3, from 0.8 to 1.6).

\begin{table}[t]
\centering
\caption{\textbf{Intervals for the main results.} 95\% bootstrap intervals over the 980 changes for the main results.}
\label{tab:ci_main}
\small
\setlength{\tabcolsep}{2.4pt}
\begin{tabular}{@{}l|ccc@{}}
\toprule
 & Judged $\approx$ & Cause $\downarrow$ & Gap $\downarrow$ \\
\midrule
Plain prompt & 16.0--18.4 & 75.9--78.2 & 99.5--103.2 \\
+ Norm card & 79.3--80.7 & 9.1--10.3 & 87.8--89.9 \\
+ Norm card with rate & 63.8--65.6 & 6.1--7.1 & 69.6--71.7 \\
+ Calibrated instruction & 19.7--21.7 & 6.1--7.0 & 25.4--28.0 \\
+ Style instruction & 0.1--0.2 & 4.9--5.8 & 9.2--10.3 \\
+ Faithful instruction & 0.3--0.4 & 0.3--1.1 & 2.0--2.8 \\
+ Balanced retrieval & 31.0--34.8 & 29.4--32.3 & 76.3--81.3 \\
+ Fixed sentences & 4.3--5.3 & 12.9--14.2 & 16.8--19.0 \\
+ Random-size sentences & 5.7--7.1 & 19.2--21.5 & 25.3--28.6 \\
+ Rule deleting all words & 0.0--0.0 & 75.9--78.2 & 85.7--88.5 \\
DRG & 1.3--2.1 & 76.7--79.1 & 84.1--86.9 \\
Template & 0.0--0.0 & 0.2--1.1 & 1.7--2.6 \\
Self-Refine & 8.6--10.0 & 48.3--50.4 & 65.7--68.9 \\
CoVe & 5.8--7.2 & 6.3--7.2 & 11.0--12.9 \\
FAVA & 15.8--18.2 & 74.6--76.9 & 97.8--101.6 \\
MiniCheck+rewrite & 5.7--6.7 & 5.8--6.9 & 10.9--12.6 \\
\ourrow \ours{} & 0.8--1.3 & 0.2--0.7 & 0.7--1.4 \\
\bottomrule
\end{tabular}
\end{table}

\noindent\textbf{Trivial rules.}
Table~\ref{tab:trivial} applies simple rules to the plain drafts and removes the components of \ours{} one at a time. Deleting every magnitude word and cutting every cause reaches a judgment rate of zero without retrieval, while it leaves more causes than \ours{}. Cutting causes accounts for about half of the drop from the plain rate, as many words sit inside the causes. Within the rule edit, a fixed threshold at 10\% gives the same result as the per-word ranges, as few words remain after retrieval and the cut. The usage norms thus matter for measuring the judgment rate and judging the remaining words, rather than for the edit.

\begin{table*}[t]
\centering
\caption{\textbf{Trivial rules and components of \ours{}.} Shares of drafts (\%) over the 17 LLMs with 95\% intervals, and the distance from the writers' word mix (JS, EMD, and JS-mix; $\times$100) with its sampling floor.}
\label{tab:trivial}
\small
\setlength{\tabcolsep}{1.6pt}
\begin{tabular}{@{}lccccccccc@{}}
\toprule
 & Judged $\approx$ & Intense & Misfit $\downarrow$ & Cause $\downarrow$ & Number $\uparrow$ & Placeholder $\downarrow$ & JS $\downarrow$ & EMD $\downarrow$ & JS-mix $\downarrow$ \\
\midrule
\multicolumn{10}{l}{\textit{Rules applied to plain (P0) drafts}} \\
Plain prompt & 17.2 [16.4, 17.9] & 12.1 & 0.4 & 77.0 & 98.0 & 8.5 & 14.91 & 39.1 & \best{21} \\
(a) Rule deleting all words & 0.0 [0.0, 0.0] & 0.0 & \best{0.0} & 77.0 & 98.0 & 8.5 & 1.40 & 2.6 & -- \\
(b) Cut cause only & 8.3 [7.8, 8.8] & 7.0 & \best{0.0} & \second{2.8} & 97.7 & \second{0.9} & 6.60 & 20.8 & 25 \\
(c) Rule deleting all + cut & 0.0 [0.0, 0.0] & 0.0 & \best{0.0} & \second{2.8} & 97.7 & \second{0.9} & 1.47 & 2.7 & -- \\
(d) Fixed-10\% edit & 16.8 [16.1, 17.6] & 12.1 & 0.2 & 77.0 & 98.0 & 8.5 & 14.84 & 38.8 & 28 \\
(d') Fixed-10\% edit + cut & 8.2 [7.7, 8.7] & 7.0 & \best{0.0} & \second{2.8} & 97.7 & \second{0.9} & 6.64 & 20.9 & 28 \\
(e) Word edit & 16.8 [16.1, 17.6] & 11.9 & \best{0.0} & 77.0 & 98.0 & 8.5 & 14.96 & 38.5 & 31 \\
(e') Word edit + cut & 8.2 [7.7, 8.7] & 7.0 & \best{0.0} & \second{2.8} & 97.7 & \second{0.9} & 6.67 & 20.9 & 35 \\
\midrule
\multicolumn{10}{l}{\textit{Components of \ours{} (on size-matched retrieval drafts)}} \\
Retrieval (retr.) only & 4.0 [3.6, 4.5] & 3.4 & \second{0.1} & 20.9 & \best{99.9} & 1.3 & 2.34 & 9.0 & \second{23} \\
Retr. + cut & \best{1.1 [0.8, 1.3]} & 0.9 & \best{0.0} & \best{0.4} & \second{99.8} & \best{0.2} & \best{0.82} & \best{2.3} & 35 \\
Retr. + word edit & 4.0 [3.6, 4.4] & 3.3 & \best{0.0} & 20.9 & \best{99.9} & 1.3 & 2.44 & 9.1 & 34 \\
\ourrow Full (\ours{}) & \second{1.0 [0.8, 1.2]} & 0.9 & \best{0.0} & \best{0.4} & \second{99.8} & \best{0.2} & 0.87 & \second{2.4} & 42 \\
Retr. + cut + fixed-10\% edit & \second{1.0 [0.8, 1.3]} & 0.9 & \best{0.0} & \best{0.4} & \second{99.8} & \best{0.2} & \second{0.84} & \best{2.3} & 28 \\
Retr. + cut + delete all & 0.0 [0.0, 0.0] & 0.0 & \best{0.0} & \best{0.4} & \second{99.8} & \best{0.2} & 1.47 & 2.7 & -- \\
\ourrow \ours{} (style) & 0.0 [0.0, 0.0] & 0.0 & \best{0.0} & \best{0.4} & \second{99.8} & 2.7 & 1.39 & 2.6 & -- \\
\specialrule{0.4pt}{1pt}{1pt}\specialrule{0.4pt}{0pt}{1pt}
\writerrow 10-K writers, same 980 changes & 1.5 [0.8, 2.3] & 0.4 & 0.4 & -- & 75.1 & 0.0 & 0.66 & 1.9 & -- \\
Sampling floor & -- & -- & -- & -- & -- & -- & 0.02 & 0.3 & 1 \\
\bottomrule
\end{tabular}
\end{table*}

\begin{table*}[t]
\centering
\caption{\textbf{Distance to the writers' word mix.} Judged (\%, mean of 7 size bins), mild words below 2\% and intense words above 80\% (\%), and the distance from the writers' per-bin word mix ($\times$100).}
\label{tab:dist}
\small
\setlength{\tabcolsep}{2.5pt}
\begin{tabular}{@{}lcccccc@{}}
\toprule
 & Judged $\approx$ & Mild $<$2 & Intense $>$80 & JS $\downarrow$ & EMD $\downarrow$ & JS-mix $\downarrow$ \\
\midrule
Plain prompt & 17.2 & 25.7 & 47.9 & 14.91 [13.98, 15.98] & 39.1 & 21 \\
+ Calibrated instruction & 20.7 & 15.4 & 50.5 & 18.16 [17.52, 18.88] & 48.7 & \best{11} \\
+ Size-matched retrieval & 4.0 & \second{3.6} & 14.0 & 2.34 [1.99, 2.81] & 9.0 & 23 \\
+ Style instruction & 0.1 & 0.2 & \best{0.5} & 1.09 [0.96, 1.25] & \best{2.3} & -- \\
+ Rule edit & 8.2 & 8.2 & 31.8 & 6.67 [6.15, 7.27] & 20.9 & 35 \\
CoVe & 6.5 & \best{3.1} & 27.2 & 5.12 [4.73, 5.56] & 16.8 & 27 \\
MiniCheck+rewrite & 6.2 & 9.1 & 18.5 & 3.88 [3.54, 4.29] & 13.5 & 24 \\
DRG & \best{1.6} & 1.1 & 4.2 & \best{0.69 [0.56, 1.14]} & 3.0 & \second{17} \\
\midrule
Rule deleting all words & 0.0 & 0.1 & \second{0.0} & 1.40 [1.25, 1.55] & 2.6 & -- \\
Rule deleting all words + cut & 0.0 & 0.0 & \second{0.0} & 1.47 [1.34, 1.62] & 2.7 & -- \\
Retrieval + cut + fixed-10\% edit & \second{1.0} & 0.9 & 3.5 & \second{0.84 [0.70, 1.06]} & \best{2.3} & 28 \\
\midrule
\ourrow \ours{} & \second{1.0} & 0.9 & 3.5 & 0.87 [0.73, 1.10] & \second{2.4} & 42 \\
\ourrow \ours{} (style) & 0.0 & 0.1 & \second{0.0} & 1.39 [1.25, 1.55] & 2.6 & -- \\
\specialrule{0.4pt}{1pt}{1pt}\specialrule{0.4pt}{0pt}{1pt}
\writerrow 10-K writers, 28K test-firm changes & 1.5 & 2.9 & 1.0 & 0 & 0 & 0 \\
\writerrow 10-K writers, same 980 changes & 1.5 & 5.0 & 0.7 & 0.66 [0.60, 1.42] & 1.9 & -- \\
Sampling floor at the drafts' $n$ & -- & -- & -- & 0.02 & 0.3 & 1 \\
\bottomrule
\end{tabular}
\end{table*}

\setcounter{table}{25}
\begin{table*}[t]
\centering
\caption{\textbf{Drafting results for each model.} Share of drafts (\%) with a magnitude word (Judged) and with a word that \ours{} flags (Misfit), for the plain prompt, the norm card, size-matched retrieval alone, and \ours{}.}
\label{tab:permodel}
\small
\begin{tabular}{@{}l|ccc>{\columncolor{yellow!15}}c|ccc>{\columncolor{yellow!15}}c@{}}
\toprule
 & \multicolumn{4}{c|}{Judged $\approx$} & \multicolumn{4}{c}{Misfit $\downarrow$} \\
\cmidrule(lr){2-5}\cmidrule(lr){6-9}
 & Plain & Norm card & Retrieval & \ours{} & Plain & Norm card & Retrieval & \ours{} \\
\midrule
Qwen2.5-7B & \second{4.1} & 96.5 & 5.1 & \best{0.4} & \second{0.1} & 27.6 & \best{0.0} & \best{0.0} \\
Qwen2.5-14B & 19.6 & 92.3 & \second{3.2} & \best{1.1} & 0.2 & 17.0 & \second{0.1} & \best{0.0} \\
Qwen2.5-32B & 25.9 & 87.2 & \second{3.9} & \best{0.4} & 0.3 & 12.8 & \second{0.1} & \best{0.0} \\
Qwen2.5-72B & 30.2 & 93.0 & \second{9.5} & \best{1.2} & 1.0 & 14.9 & \second{0.3} & \best{0.0} \\
Qwen3-8B & \second{9.4} & 67.7 & \second{9.4} & \best{0.3} & \second{0.2} & 15.3 & \second{0.2} & \best{0.0} \\
Qwen3-14B & \second{2.9} & 29.5 & \best{0.6} & \second{0.1} & \best{0.0} & \second{4.1} & \best{0.0} & \best{0.0} \\
Qwen3-32B & \best{2.0} & 73.2 & \second{0.4} & 0.0 & \second{0.1} & 21.8 & \best{0.0} & \best{0.0} \\
Llama-3.1-8B & 16.3 & 86.9 & \second{3.8} & \best{1.2} & 0.8 & 24.1 & \second{0.1} & \best{0.0} \\
Llama-3.1-70B & 12.8 & 97.3 & \best{0.4} & \second{0.2} & \second{0.2} & 19.4 & \best{0.0} & \best{0.0} \\
Llama-3.3-70B & 34.8 & 96.8 & \second{2.8} & \best{1.2} & 0.8 & 20.8 & \second{0.1} & \best{0.0} \\
Mistral-7B & 35.1 & 53.0 & \second{4.1} & \best{2.1} & 0.6 & 3.0 & \second{0.1} & \best{0.0} \\
Mistral-Small-24B & 16.4 & 94.6 & \second{3.0} & \best{1.4} & \second{0.2} & 22.9 & \second{0.2} & \best{0.0} \\
Gemma-2-9B & 9.8 & 54.9 & \second{5.6} & \best{4.7} & \best{0.0} & \second{7.9} & \best{0.0} & \best{0.0} \\
Gemma-2-27B & \second{19.0} & 99.5 & \best{1.6} & \best{1.4} & \best{0.0} & \second{17.2} & \best{0.0} & \best{0.0} \\
Gemma-3-12B & 31.4 & 72.2 & \best{1.2} & \second{0.8} & \second{0.5} & 3.3 & \best{0.0} & \best{0.0} \\
Phi-4 & \second{8.2} & 93.4 & 13.8 & \best{0.8} & \second{0.1} & 28.6 & 0.2 & \best{0.0} \\
gpt-oss-20B & 13.9 & 71.6 & \best{0.4} & \second{0.0} & 0.9 & 8.8 & \second{0.1} & \best{0.0} \\
\bottomrule
\end{tabular}
\end{table*}

\setcounter{table}{24}
\begin{table}[t]
\centering
\caption{\textbf{Shifts in LLM reader judgments.} Shift in what eight LLM readers rate on 1 to 7 scales and texts read as giving a reason (\%), against the number only. Marks show the smallest shifts from the number.}
\label{tab:reader}
\small
\setlength{\tabcolsep}{1.6pt}
\begin{tabular}{@{}l|cccc@{}}
\toprule
Added to the number & Size & Importance & Sentiment & Reason \\
\midrule
Inflating misfit word & +0.29 & +0.20 & $-$0.03 & 0 \\
Deflating misfit word & $-$0.50 & $-$0.52 & $-$0.14 & 0 \\
Invented cause & +0.10 & +0.12 & $-$0.06 & 99 \\
\midrule
Plain prompt & \second{+0.14} & +0.33 & +0.31 & 73 \\
\ \ + Rule edit & \best{+0.01} & \second{+0.05} & \best{+0.00} & \second{4} \\
\ourrow \ours{} & \best{$-$0.01} & \best{+0.01} & \second{$-$0.01} & \best{1} \\
\specialrule{0.4pt}{1pt}{1pt}\specialrule{0.4pt}{0pt}{1pt}
\writerrow 10-K writers & $-$0.15 & +0.06 & +0.08 & 29 \\
\bottomrule
\end{tabular}
\end{table}

\setcounter{table}{26}

\noindent\textbf{Rate and mix.}
A judgment rate of zero is lower than the writers' rate, and the rate alone cannot rank such outputs. Table~\ref{tab:dist} therefore compares, for each size bin, the share of texts with no, mild, moderate, or intense words with that of the writers. JS is the Jensen-Shannon divergence and EMD the earth mover's distance over these four ordered classes, averaged over the 7 bins, and JS-mix uses only the texts with a judgment word. The sampling floor is the value expected when the texts of a condition are drawn from the writers' own words at each size. \ours{} and DRG come closest to the writers, while deleting every word and \ours{} (style) remove the mild words that writers use on small changes. DRG comes slightly closer on JS (0.69 vs.\ 0.87, with overlapping intervals) but keeps causes in 77.9\% of its drafts. JS-mix is highest for \ours{} (42), as its few remaining words are mostly intense words on large changes. No condition reaches the sampling floor.

\section{Simulated Reader Study}
\label{app:reader}

\noindent\textbf{Design.}
We sample 600 changes from the drafting items, about 86 per size bin, and build up to 11 versions of each that state the same number. They add a fitting word, a misfit word, the word the drafting LLMs used most, or an invented cause, and they include the plain draft, its rule edit, the \ours{} draft, and the writer's sentence. Each version is shown once with its number and once with the number replaced by \textit{X}, giving 12K texts.

\noindent\textbf{Readers.}
Llama-3.3-70B, Qwen2.5-72B, Mistral-Small-24B, Gemma-2-27B, Qwen3-32B, Phi-4, Qwen2.5-7B, and Llama-3.1-8B read each text in a separate call, blind to the version, and rate the size, importance, sentiment, and confidence on 1 to 7 scales. They also name the reason the text gives, if any. Effects are differences within items, averaged over readers, with bootstrap intervals over items. Readers agree closely on size, importance, and sentiment (mean pairwise Spearman correlation of 0.93, 0.81, and 0.89) but not on confidence (0.26), which we therefore omit from the tables.

\noindent\textbf{Results.}
As Table~\ref{tab:reader} shows, an inflating misfit word raises the rated size by 0.29 and a deflating one lowers it by 0.50, and readers report an invented cause as the reason for 99\% of the texts. Plain drafts read as more important (+0.33) and more positive (+0.31) than the number alone, while \ours{} removes almost all of this shift.

\noindent\textbf{Number hidden.}
Without the number, the wording alone tells how large the change is. A fitting word lets readers recover the size far better than the number-only sentence, while a misfit word reverses the relation. The LLMs' own words mostly fit the size, so the harm sits in the misfit minority, invented causes, and the positive slant of plain drafts. A financial sentiment classifier~\citep{huang2023finbert} shows the same slant, which the edit removes.

\section{Further Analyses}
\label{app:results}

\subsection{Writers and models}
\label{app:wm}

\noindent\textbf{Judgment rate of the writers.}
As the 980 drafted changes contain few writer sentences with a judgment word, we also draw 2{,}000 changes of the test firms for each size bin and direction, measured with the same lexicon. They give the same judgment rate as the drafted changes, and Section~\ref{sec:res_audit} reports their rates by the size of the change.

\begin{table*}[t]
\centering
\caption{\textbf{Ablation of retrieval.} Shares of drafts (\%) averaged over the 17 LLMs. Copied drafts share eight consecutive words with a retrieved sentence, and Style adds that filings rarely use magnitude words.}
\label{tab:ablation}
\small
\setlength{\tabcolsep}{3.5pt}
\begin{tabular}{@{}l|ccccccc@{}}
\toprule
 & Judged $\approx$ & Intense & Number $\uparrow$ & Cause $\downarrow$ & Placeholder $\downarrow$ & Invented $\downarrow$ & Copied $\downarrow$ \\
\midrule
Plain prompt & 17.2 & 12.1 & 98.0 & 77.0 & 8.5 & 0.9 & -- \\
+ Style instruction & \best{0.1} & \best{0.1} & \second{99.8} & \second{5.3} & 3.1 & 2.3 & -- \\
\midrule
+ Random-size sentences & 6.3 & 5.5 & 99.7 & 20.3 & 1.7 & 0.5 & \best{0.1} \\
+ Size-matched, 2 sentences & 5.4 & 4.4 & \second{99.8} & 22.7 & 1.6 & \second{0.3} & \second{0.4} \\
+ Size-matched, 6 sentences & 4.0 & 3.4 & \best{99.9} & 20.9 & 1.3 & \best{0.1} & 0.6 \\
+ Size-matched, 12 sentences & 3.2 & 2.7 & \second{99.8} & 19.3 & \second{0.9} & \best{0.1} & 0.7 \\
\midrule
+ Size-matched retrieval + style & \second{0.0} & \second{0.0} & \best{99.9} & \best{1.1} & \best{0.1} & \best{0.1} & 0.5 \\
\bottomrule
\end{tabular}
\end{table*}

\noindent\textbf{Natural mix of sizes.}
The drafting items have the same number of changes in each size bin, which gives more weight to large changes than the filings do. We therefore weight the bins by the share of changes of each size in the test firms. With these weights, plain drafts judge far more often than the writers under each instruction P0, P1, and P2.

\noindent\textbf{Results for each model.}
Table~\ref{tab:permodel} reports the drafting results for each model. The norm card raises the rate for every model, while size-matched retrieval lowers or keeps it for 15 of the 17 models and raises it only for Phi-4 and Qwen2.5-7B.

\subsection{Retrieval design}
\label{app:retdesign}

\noindent\textbf{Number of retrieved items.}
Table~\ref{tab:ablation} varies the retrieved sentences. More size-matched sentences lower the judgment rate further, and the share of drafts that copy eight consecutive words of a retrieved sentence stays small. The firm split does not remove text that two firms file under different identifiers, and 59 of the 980 test sentences (6\%) have an exact copy in the pool. Removing all near-duplicates found with sentence embeddings barely changes the judgment rate of \ours{}.

\begin{table}[t]
\centering
\caption{\textbf{Checks against tuning and memorization.} Top: Drafts (\%) of three LLMs with unrestricted or restricted retrieval. Bottom: Completions of test and recent sentences by four LLMs.}
\label{tab:val}
\small
\setlength{\tabcolsep}{8.0pt}
\begin{tabular}{@{}l|cc|cc@{}}
\toprule
 & \multicolumn{2}{c|}{Invented} & \multicolumn{2}{c}{Placeholder} \\
\cmidrule(lr){2-3}\cmidrule(lr){4-5}
 & Dev $\downarrow$ & Test $\downarrow$ & Dev $\downarrow$ & Test $\downarrow$ \\
\midrule
Unrestricted & 34.5 & 31.1 & 14.8 & 15.7 \\
Restricted & \best{0.8} & \best{0.8} & \best{0.6} & \best{0.7} \\
\midrule
 & \multicolumn{2}{c|}{Exact match} & \multicolumn{2}{c}{ROUGE-L} \\
 & Test & 2025 & Test & 2025 \\
\midrule
Completion & 0.0 & 0.0 & 24.2 & 24.2 \\
\bottomrule
\end{tabular}
\end{table}

\begin{table}[t]
\centering
\caption{\textbf{Cost after drafting.} Extra model calls, generated tokens per draft, and GPU seconds per 1,000 drafts, for the editors given our feedback (Appendix~\ref{app:prior}).}
\label{tab:cost}
\small
\setlength{\tabcolsep}{5.6pt}
\begin{tabular}{@{}l|ccc@{}}
\toprule
 & Calls $\downarrow$ & Tokens $\downarrow$ & GPU seconds $\downarrow$ \\
\midrule
Self-Refine + Style & 3.4 & 307 & 451 \\
CoVe + Style & 5.0 & 194 & 232 \\
DRG + Flags & \second{0.8} & \second{25} & \second{68} \\
\midrule
\ourrow \ours{} & \best{0} & \best{0} & \best{0} \\
\ourrow \ours{} (style) & \best{0} & \best{0} & \best{0} \\
\bottomrule
\end{tabular}
\end{table}

\noindent\textbf{Paired comparisons.}
To compare retrieval strategies, we bootstrap the 980 changes with the models paired. Size-matched sentences give a lower judgment rate than fixed or random-size sentences, and six sentences lower than two, with 95\% bootstrap intervals of the differences above zero.

\noindent\textbf{Retrieval with amounts.}
Our first version of size-matched retrieval took the six closest 10-K sentences as they are. Many state dollar amounts, such as \textit{``to \$10.7 million from \$10.5 million''}, which the input lacks. With these sentences, the models copied the pattern and often invented amounts or inserted placeholders such as \textit{``\$X million''}.

\noindent\textbf{Restricted retrieval.}
We therefore restrict the retrieved sentences to those that state the change as a percentage without dollar amounts or causes. As this change was made after seeing the drafts of the same items, we report the first version here as well and repeat the comparison on 490 development changes of the reference firms, excluding sentences of the same firm from the pool (Table~\ref{tab:val}). The restriction removes invented amounts and placeholders by as much on these changes as on the test changes. Examples in a prompt shape not only the wording but also the facts of a draft.

\noindent\textbf{Memorization.}
EDGAR filings are likely in the pretraining data. We give four LLMs the start of 300 test-firm writer sentences, cut before the first number, and of 300 sentences of similar length filed in 2025, and let them complete each greedily. No completion reproduces a sentence's rest, and the overlap with the true rest is equal for both sets. In plain drafts, an 8-gram shared with the writer almost always repeats the input's line item.

\subsection{Cost}
\label{app:cost}

\noindent\textbf{Cost of \ours{}.}
Table~\ref{tab:cost} reports the cost of each option, counted with the tokenizer of Qwen2.5-7B. The check, the lookup of the retrieved sentences, and the rule edit run on a CPU without a model call. Before drafting, retrieval adds 173 prompt tokens per draft (from 97 to 270) and the style sentence adds 29, while each editor call also re-reads the input and the draft. The revision calls the model again only for the flagged drafts.

\begin{table*}[t]
\centering
\caption{\textbf{Paragraphs from a table of four changes.} Magnitude and misfit words per 100 changes, the largest share of one LLM (Max), and paragraphs with the numbers, a cause, or a placeholder (\%). Writers: their four sentences.}
\label{tab:paragraph}
\small
\setlength{\tabcolsep}{2.5pt}
\begin{tabular}{@{}l|cccccc@{}}
\toprule
 & Words $\approx$ & Max & Misfit $\downarrow$ & Number $\uparrow$ & Cause $\downarrow$ & Placeholder $\downarrow$ \\
\midrule
Plain prompt & 27.6 & 40.4 & 2.8 & 99.3 & 98.8 & 6.9 \\
+ Rule edit & 16.4 & 25.8 & \best{0.0} & 96.2 & 15.8 & \second{1.7} \\
+ Size-matched retrieval & 31.9 & 46.2 & 3.3 & \second{99.5} & 97.8 & 9.3 \\
\ourrow \ours{} & 18.4 & 29.2 & \best{0.0} & 98.1 & \best{9.8} & \second{1.7} \\
+ Style instruction & \second{3.9} & \second{13.7} & \second{0.3} & \best{99.6} & 77.3 & 3.0 \\
\ourrow \ours{} (style) & \best{2.1} & \best{9.6} & \best{0.0} & 97.2 & \second{12.9} & \best{1.5} \\
\specialrule{0.4pt}{1pt}{1pt}\specialrule{0.4pt}{0pt}{1pt}
\writerrow 10-K writers & 1.6 & - & 0.4 & 100.0 & 60.8 & 0.0 \\
\bottomrule
\end{tabular}
\end{table*}

\begin{table}[t]
\centering
\caption{\textbf{Independent check of the rule edit.} Drafts (\%) in which each judge (Llama-3.3-70B, Mistral-Small-24B, Qwen2.5-72B) finds a cause or a complete sentence; 400 drafts per row and 337 for writers.}
\label{tab:judge_edit}
\small
\setlength{\tabcolsep}{2.5pt}
\begin{tabular}{@{}l|ccc|ccc@{}}
\toprule
 & \multicolumn{3}{c|}{Cause $\downarrow$} & \multicolumn{3}{c}{Complete $\uparrow$} \\
\cmidrule(lr){2-4}\cmidrule(lr){5-7}
 & Llama & Mistral & Qwen & Llama & Mistral & Qwen \\
\midrule
Plain prompt & 76.5 & 71.2 & 72.0 & 91.2 & \second{99.0} & 98.2 \\
+ Rule edit & \second{1.5} & \second{1.5} & \second{1.5} & 97.2 & 97.5 & 98.8 \\
+ Retrieval & 29.2 & 29.2 & 28.2 & 97.2 & \second{99.0} & 99.2 \\
\ourrow \ours{} & \best{1.2} & \best{1.0} & \best{0.8} & \second{98.8} & 98.5 & \second{99.8} \\
+ Style & 8.0 & 6.8 & 7.0 & 96.2 & 98.8 & 99.2 \\
\midrule
Self-Refine & 49.2 & 49.5 & 49.8 & 92.5 & 97.8 & 97.5 \\
CoVe & 6.8 & 7.5 & 7.8 & \best{99.0} & \best{99.2} & \best{100.0} \\
FAVA & 75.2 & 70.2 & 70.8 & 89.8 & 98.2 & 97.0 \\
\begin{tabular}[t]{@{}l@{}}MiniCheck+\\rewrite\end{tabular} & 7.5 & 8.2 & 8.5 & 98.0 & 98.2 & \second{99.8} \\
DRG & 79.0 & 74.0 & 74.2 & 91.5 & \second{99.0} & 98.2 \\
\specialrule{0.4pt}{1pt}{1pt}\specialrule{0.4pt}{0pt}{1pt}
\writerrow 10-K writers & 31.5 & 32.6 & 31.2 & 92.6 & 92.3 & 97.3 \\
\bottomrule
\end{tabular}
\end{table}

\subsection{Paragraphs}
\label{app:para}

\noindent\textbf{Paragraphs from a table.}
For paragraphs, we take 400 filings of the test firms that report at least four changes of plainly named items as percentages in separate sentences, and give four as a table. Retrieval adds the two closest sentences of the reference firms for each change. In paragraphs, the rule edit also drops sentences stating none of the table's percentages, such as an opening or closing remark, before cutting causes and misfit words.

\noindent\textbf{Results on paragraphs.}
As shown in Table~\ref{tab:paragraph}, plain paragraphs carry many more magnitude words per change than single sentences, often in such remarks (e.g., \textit{``we observed a significant decline in the prices of key commodities''}). Retrieved sentences raise the rate, as models compare changes with another firm's (e.g., \textit{``echoing the decrease in time charter revenue for a competitor''}). The style instruction and rule edit both lower it, and together come closest to the writers.

\noindent\textbf{Paragraphs from a table and a note.}
For the setting closer to deployment, we take 300 paragraphs of the test firms that report three to five changes and give a cause for at least one. Each LLM receives a table of the changes and a note that lists the writer's own cause clauses with magnitude words removed, and writes one paragraph. Three LLM judges, gpt-oss-20B, Mistral-Small-24B, and Qwen2.5-32B, mark each cause as supported by the note or not, and whether each noted cause is kept. The writers' 149 paragraphs reach 28.9\% with an unsupported cause, as they also explain items whose cause the note does not list, and all three LLM judges agree on only 17 of these 43 flagged paragraphs.

\subsection{Checks of the rule edit}
\label{app:ruleedit}

\noindent\textbf{Open-vocabulary check.}
Judged and Cause share the lexicon and connectives of \ours{}. We therefore also give three LLM judges, gpt-oss-20B, Mistral-Small-24B, and Qwen2.5-32B, the input and one text, and ask whether the text uses any word or phrase that judges the size of the change, including evaluative words such as \textit{robust}, and whether it adds a reason or interpretation that the input does not give. We sample the same 500 changes and drafts for each method and take the majority vote. The judges agree closely (Fleiss $\kappa$ of 0.96 and 0.94), and the main results report the majority vote in the Judg.\ and Expl.\ columns.

\setcounter{table}{32}
\begin{table*}[t]
\centering
\caption{\textbf{Robustness of the check.} (a)~Norms from two sets of filings: rank correlation of word medians ($\rho$), and agreement (\%) and Cohen's $\kappa$ of the flag on 79K drafts; halves and subsets average 20 draws, and -- marks fewer than three words with 30 uses. (b)~Range of a fit: detection on clear cases and flagged texts with a word (\%).}
\label{tab:stab}
\small
\begin{minipage}[t]{0.46\textwidth}
\centering
\setlength{\tabcolsep}{4.0pt}
\begin{tabular}[t]{@{}l|ccc@{}}
\multicolumn{4}{@{}l}{\textit{(a) Norms A vs.\ norms B}} \\
\toprule
 & $\rho$ $\uparrow$ & Agree $\uparrow$ & $\kappa$ $\uparrow$ \\
\midrule
1993--2005 vs.\ 2006--2020 & 0.94 & 96.4 & 0.82 \\
2006--2020 vs.\ 2016--2025 & 1.00 & 98.7 & 0.93 \\
Random halves of firms & 0.96 & 95.4 & 0.78 \\
Small vs.\ large firms & 0.83 & 94.1 & 0.72 \\
Bank-like vs.\ other firms & 1.00 & 92.2 & 0.59 \\
\midrule
Full vs.\ 50\% of firms & 0.98 & 93.8 & 0.73 \\
Full vs.\ 25\% of firms & 0.96 & 92.7 & 0.69 \\
Full vs.\ 10\% of firms & -- & 92.4 & 0.67 \\
\bottomrule
\end{tabular}
\end{minipage}\hfill
\begin{minipage}[t]{0.52\textwidth}
\centering
\setlength{\tabcolsep}{3.0pt}
\begin{tabular}[t]{@{}l|cccc|cc@{}}
\multicolumn{7}{@{}l}{\textit{(b) Range of a fit}} \\
\toprule
Range & Prec.\ $\uparrow$ & Rec.\ $\uparrow$ & F1 $\uparrow$ & False $\downarrow$ & Drafts & Summ. \\
\midrule
$q_{05}$--$q_{95}$ & \best{0.906} & 1.000 & \best{0.950} & \best{0.083} & 0.7 & 0.6 \\
\ourrow $q_{10}$--$q_{90}$ & \second{0.868} & 1.000 & \second{0.929} & \second{0.121} & 2.1 & 1.0 \\
$q_{25}$--$q_{75}$ & 0.724 & 1.000 & 0.840 & 0.302 & 8.9 & 5.6 \\
\bottomrule
\end{tabular}
\end{minipage}
\end{table*}

\setcounter{table}{31}
\begin{table}[t]
\centering
\caption{\textbf{Revising the flagged drafts.} Shares of drafts (\%) averaged over the 17 LLMs. Only the drafts with a misfit word are revised once by the model or edited by rule, without the cut of causes, and the rest are kept.}
\label{tab:revise}
\small
\setlength{\tabcolsep}{1.0pt}
\begin{tabular}{@{}l|ccccc@{}}
\toprule
 & Judged $\approx$ & Intense & Misfit $\downarrow$ & Number $\uparrow$ & Dir. $\uparrow$ \\
\midrule
Plain prompt & 17.2 & 12.1 & \second{0.4} & 98.0 & 99.6 \\
+ Model revision & \second{16.9} & \second{12.0} & \best{0.0} & 98.0 & 99.6 \\
+ Word edit & \best{16.8} & \best{11.9} & \best{0.0} & 98.0 & 99.6 \\
\midrule
Norm card & 80.0 & \second{39.0} & 15.9 & \second{88.8} & \best{99.7} \\
+ Model revision & \second{78.1} & 42.3 & \second{9.8} & \best{89.2} & \second{99.6} \\
+ Word edit & \best{64.3} & \best{34.3} & \best{0.1} & \second{88.8} & \best{99.7} \\
\midrule
Calibrated & 20.7 & \second{15.2} & 1.8 & 99.0 & 99.7 \\
+ Model revision & \second{19.8} & 15.4 & \second{0.5} & 99.0 & 99.7 \\
+ Word edit & \best{18.9} & \best{14.6} & \best{0.0} & 99.0 & 99.7 \\
\bottomrule
\end{tabular}
\end{table}

\setcounter{table}{33}

\noindent\textbf{Independent rule-edit check.}
We also ask three judges whether each sentence gives a cause for the change and whether it is a complete sentence that reports the change. The judges are Llama-3.3-70B, Mistral-Small-24B, and Qwen2.5-72B. They score 400 drafts in each condition, sampled from the 10-model subset, and the writers' sentences for the same changes as a reference. Two of the three judges, Llama-3.3-70B and Mistral-Small-24B, wrote none of these drafts themselves.

As shown in Table~\ref{tab:judge_edit}, all three judges find that the rule edit removes almost all causes. The edited drafts remain complete sentences as often as the drafts before the rule edit and the writers' own sentences. The judges also confirm that CoVe and MiniCheck+rewrite remove most causes, though fewer than \ours{}, while Self-Refine, FAVA, and DRG keep many of these causes.

\noindent\textbf{Editing flagged drafts.}
Table~\ref{tab:revise} reports the rule edit and revision on plain drafts and on drafts of the norm card and the calibrated instruction. Revision by the model still leaves misfit words in 9.8\% of the norm-card drafts, while the word edit removes almost all of them, and neither of the two changes the stated number or the direction of the change.

\subsection{Robustness of the check}
\label{app:sensitivity}

\noindent\textbf{Other filings and ranges.}
Table~\ref{tab:stab} compares norms built from two disjoint sets of filings, such as different periods, firm sizes, or random halves, and repeats the check with the 5th to 95th and the 25th to 75th percentiles. Norms from other filings flag almost the same drafts. A wider range lowers the false flags on writers' sentences without a loss in recall, while a narrower range flags many of the writers' own sentences. The share of flagged drafts stays small for default and wider ranges, so the finding that plain drafts rarely misfit does not depend on this choice of percentiles.

\subsection{Word choice of the models}
\label{app:reading}

\noindent\textbf{Where drafts misfit.}
Plain drafts put most of their words where a word rarely misfits: Intense words on changes above 20\%, which are almost never flagged, and mild words on changes below 5\%. Among the few intense words on changes between 10\% and 20\%, where writers switch from mild to intense words, most are flagged by the norms.

\begin{table}[t]
\centering
\caption{\textbf{Forced word choice in 10-K sentences.} Over: mild-word blanks filled with an intense word (rate). Thr.: the size (\%) where the model switches to intense words.}
\label{tab:choose}
\small
\setlength{\tabcolsep}{1.8pt}
\begin{tabular}{@{}l|ccc|ccc@{}}
\toprule
 & \multicolumn{3}{c|}{Plain prompt} & \multicolumn{3}{c}{+ Retrieval} \\
\cmidrule(lr){2-4}\cmidrule(lr){5-7}
 & Accuracy & Over & Thr. & Accuracy & Over & Thr. \\
\midrule
Qwen2.5-7B & \second{0.351} & 0.224 & 5.8 & 0.521 & \second{0.039} & \second{15.8} \\
Qwen2.5-14B & 0.269 & 0.286 & 3.9 & 0.478 & 0.059 & 9.8 \\
Qwen2.5-72B & \best{0.370} & \second{0.020} & \second{18.7} & 0.552 & 0.011 & \best{14.6} \\
Qwen3-8B & 0.272 & 0.728 & 0.4 & \best{0.570} & 0.098 & 8.9 \\
Llama-3.1-8B & 0.219 & 0.541 & 3.9 & 0.326 & 0.120 & 33.8 \\
Llama-3.1-70B & 0.281 & 0.008 & 24.3 & \second{0.566} & 0.025 & 13.2 \\
Mistral-7B & 0.200 & 0.910 & 0.0 & 0.346 & 0.317 & 5.6 \\
Gemma-2-9B & 0.293 & \best{0.045} & \best{16.6} & 0.488 & 0.034 & 18.9 \\
Gemma-3-12B & 0.302 & 0.305 & 4.9 & 0.512 & \best{0.042} & 13.6 \\
Phi-4 & 0.327 & 0.272 & 4.9 & 0.509 & 0.070 & 8.6 \\
\midrule
Corpus speaker & 0.348 & 0.045 & 14.8 & -- & -- & -- \\
\bottomrule
\end{tabular}
\end{table}

\noindent\textbf{Forced choice of a word.}
We remove the writer's magnitude word from 791 test-firm 10-K sentences and ask each model to fill the blank with one of eight common words or none, against a \textit{corpus speaker} that picks the writers' most common word for each size. As shown in Table~\ref{tab:choose}, most 7B to 14B models switch from mild to intense words at much smaller changes than the writers, which explains the misfit words of the norm card, and retrieval moves most boundaries toward the writers.

\section{Detecting Misfit Words}
\label{app:detect}

\noindent\textbf{Summary.}
The misfit flag reaches an AUROC of 0.97 on swapped words and 0.81 on real drafts, close to G-Eval with Qwen2.5-72B (0.98 and 0.82), with no model call, and on injected words its F1 of 0.90 beats a fixed threshold (0.78), though flagging all words has a higher F1 on real drafts (Table~\ref{tab:detect_all}).

\begin{table}[t]
\centering
\caption{\textbf{Detecting misfit words.} AUROC (AUC) and F1 on swapped words in 10-K sentences, real drafts judged by three LLMs, and injected words. Methods are ranked within each cost group.}
\label{tab:detect_all}
\small
\setlength{\tabcolsep}{3.1pt}
\begin{tabular}{@{}l|c|cc|c@{\hspace{\tabcolsep}}}
\toprule
 & Swapped & \multicolumn{2}{c|}{Real} & Injected \\
\cmidrule(lr){2-2}\cmidrule(lr){3-4}\cmidrule(lr){5-5}
 & AUC $\uparrow$ & AUC $\uparrow$ & F1 $\uparrow$ & F1 $\uparrow$ \\
\midrule
\multicolumn{5}{l}{\textit{No model call}} \\
Flag all words & -- & 0.50 & \best{0.76} & 0.76 \\
Fixed 10\% rule & \best{0.99} & 0.76 & 0.54 & \second{0.78} \\
\ours{}, classes & -- & \second{0.78} & 0.60 & \best{0.90} \\
\ourrow \ours{} & 0.97 & \best{0.81} & \second{0.71} & \best{0.90} \\
\midrule
\multicolumn{5}{l}{\textit{Trained classifier}} \\
Exaggeration & 0.60 & 0.49 & \best{0.76} & \best{0.76} \\
Certainty & \second{0.82} & \second{0.59} & \second{0.64} & \second{0.66} \\
MiniCheck-FT5 & \best{0.87} & \best{0.62} & 0.54 & 0.39 \\
\midrule
\multicolumn{5}{l}{\textit{LLM evaluator, one call per text}} \\
G-Eval, Qwen2.5-72B & \best{0.98} & \best{0.82} & 0.38 & 0.66 \\
G-Eval, Llama-3.3-70B & 0.97 & 0.79 & \best{0.45} & \best{0.74} \\
\bottomrule
\end{tabular}
\end{table}

\noindent\textbf{Swapped words.}
We take 580 real 10-K sentences of the test firms with a mild or intense word and copy each with the word swapped for one of the other end. A swap is \textit{clear} when it puts an intense word on a change below 5\% or a mild word on one above 25\%, and a \textit{boundary} case otherwise.

We compare \ours{} with a fixed rule that flags intense words below and mild words above 10\%, and with LLM judges~\citep{zheng2023judging} of whether the word fits. We also combine each check with a judge that sees only the flagged sentences. We further compare detectors of exaggeration~\citep{wright2021semi} and certainty~\citep{pei2021measuring}, fact checkers MiniCheck, AlignScore~\citep{zha2023alignscore}, SummaC~\citep{laban2022summac}, and FActScore~\citep{min2023factscore}, and evaluators Prometheus 2~\citep{kim2024prometheus} and G-Eval~\citep{liu2023geval}, with thresholds tuned on reference firms (Appendix~\ref{app:prior}).

On clear cases, \ours{} finds every misfit but also flags some of the writers' own sentences, as defined. A fixed rule does better there, as clear cases sit on both sides of its threshold, but it cannot cover moderate words and verbs. LLM judges either flag many writers' sentences or miss misfits, depending on the model, but do better on boundary cases. Among prior detectors, only G-Eval with a large model matches \ours{}, while the others do no better than flagging every sentence or flag many of the writers' own words. Adding a judge that sees only the flagged sentences then raises the precision of the norm flag and the fixed rule.

\noindent\textbf{Real and injected errors.}
Swapped words test only the two ends of the scale, so we also test the flag on real drafts and injected words. Three open LLM judges, Mistral-Small-24B, Qwen3-32B, and Llama-3.3-70B, rate 1,000 drafts with a judgment word of the 10 LLMs as overstating, fitting, or understating the change. They are stricter than 10-K writers and agree only moderately, so we take their majority as a rough reference. The norm flag beats the fixed threshold for words of every class, while the fixed threshold does better on the 75 plain drafts alone, mostly intense words on large changes. Neither rule catches the verbs the judges object to.

We further insert 13 words that LLMs often use into plain drafts at sizes from small to very large and take the writers of held-out firms as the reference. The norms win where one threshold cannot fit every word, such as intense words on medium changes and moderate words, which a single threshold flags in the wrong direction. Ranges by class recover most of this gain, and G-Eval with a 70B model ranks real drafts about as well as the flag, although its tuned threshold flags few of them.

\noindent\textbf{Lexicon coverage.}
On 500 drafts annotated with an open vocabulary, the lexicon covers most expressions of size, and adding the missed ones barely changes the judgment rate. Evaluative words such as \textit{strong} or \textit{favorable}, which the lexicon leaves out by design, mostly sit inside unsupported causes.

\section{Baselines from Prior Work}
\label{app:prior}

\noindent\textbf{Tuning on reference firms.}
We adapt each prior detector to score one magnitude word in a sentence, given the size of the change. To tune their thresholds without the test firms, we build a development set like the test set from 662 sentences of the reference firms, giving 1{,}324 kept and swapped sentences. Each threshold, and the size cut below, maximizes F1 on the clear cases of this set.

\noindent\textbf{Exaggeration and certainty.}
The detectors of \citet{wright2021semi} and \citet{pei2021measuring} score the strength and the certainty of a claim. We score each sentence with and without its magnitude word and flag a word that adds strength below a size cut or removes it above.

\noindent\textbf{Fact checkers.}
MiniCheck-FT5 and Bespoke-MiniCheck-7B~\citep{tang2024minicheck}, AlignScore~\citep{zha2023alignscore}, and SummaC~\citep{laban2022summac} in its zero-shot form receive the input facts and the sentence without its magnitude word as the document and the full sentence as the claim. For FActScore~\citep{min2023factscore}, Qwen2.5-32B splits the sentence into atomic facts and judges each against the same document, and the score is the mean probability of \textit{``True''} over the facts.

\noindent\textbf{LLM evaluators.}
Prometheus 2~\citep{kim2024prometheus} and G-Eval~\citep{liu2023geval} with Qwen2.5-72B and Llama-3.3-70B rate on a five-point rubric whether the magnitude words fit the change, and we use the expected score as the flag.

\begin{table}[t]
\centering
\caption{\textbf{Prior editors given our feedback.} Shares of drafts (\%) of 10 LLMs when each editor also receives the style sentence (+ Style) or the flags (+ Flags) of \ours{}. All conditions, with or without our feedback, state the size in at least 95\% of drafts.}
\label{tab:fair}
\small
\setlength{\tabcolsep}{2.4pt}
\begin{tabular}{@{}l|cccc|c@{\hspace{\tabcolsep}}}
\toprule
 & Judged $\approx$ & Cause $\downarrow$ & Placeh. $\downarrow$ & Gap $\downarrow$ & Calls $\downarrow$ \\
\midrule
Self-Refine & 10.4 & 51.0 & 13.1 & 72.9 & 3.3 \\
\ \ + Style & 0.8 & 22.0 & 13.9 & 36.6 & 3.4 \\
\ \ + Flags & 11.1 & 52.4 & 15.4 & 77.3 & 3.5 \\
\midrule
CoVe & 8.4 & 9.1 & \best{0.2} & 16.2 & 5.0 \\
\ \ + Style & 0.1 & \second{2.2} & 0.5 & \second{4.1} & 5.0 \\
\ \ + Flags & 6.6 & 3.2 & \second{0.3} & 8.6 & 5.0 \\
\midrule
DRG & \second{2.0} & 80.8 & 10.0 & 91.3 & \second{0.2} \\
\ \ + Style & 0.8 & 80.3 & 10.1 & 91.1 & \second{0.2} \\
\ \ + Flags & \best{1.3} & 6.5 & 1.3 & 8.0 & 0.8 \\
\midrule
\ourrow \ours{} & \best{1.3} & \best{0.5} & \best{0.2} & \best{0.9} & \best{0} \\
\bottomrule
\end{tabular}
\end{table}

\noindent\textbf{Editors.}
All editors receive the input facts of the draft, but not the norms or flags of \ours{}. Self-Refine~\citep{madaan2023selfrefine} revises up to twice after its own feedback on accuracy, clarity, and style, and CoVe~\citep{dhuliawala2024chain} uses its factored variant with answers from the input facts alone. FAVA~\citep{mishra2024finegrained} takes the input facts as its reference, and we apply its edits. MiniCheck+rewrite sends each draft that MiniCheck-FT5 finds unsupported (below 0.5) back to the drafting model once for a rewrite.

\noindent\textbf{Delete-Retrieve-Generate.}
DRG~\citep{li2018delete} takes the strength of the wording as the attribute and our lexicon as its markers. It deletes every magnitude word, retrieves the sentence most similar in content by TF-IDF among the 300 reference-firm sentences of the same direction closest in size, and lets the drafting model rewrite the content with the magnitude words of that sentence, often none, in place of the generator that DRG trains.

\noindent\textbf{Editors given our feedback.}
Table~\ref{tab:fair} gives each editor the style sentence or the flags of \ours{} on the 10-model subset of Appendix~\ref{app:backbone}. Both help CoVe and DRG, but none comes as close to the writers as \ours{} (Gap 0.9), and the closest, CoVe with the style sentence (4.1), takes five calls.

\noindent\textbf{What the editors change.}
Self-Refine rewrites most drafts. CoVe and MiniCheck+rewrite lower the rate mostly by cutting unsupported causes. FAVA changes few drafts and marks few causes as unverifiable, although the input gives no cause. DRG rewrites only the drafts in which a magnitude word or a retrieved target marker appears.

\section{Deployment Guide}
\label{app:deploy}

\noindent\textbf{Choosing a variant.}
Every variant runs Steps 2 and 3 on the CPU after drafting, in about 0.04 seconds per 1{,}000 drafts, and differs only in what it adds to the prompt before the drafting call.
\begin{itemize}[leftmargin=*,itemsep=0pt,topsep=2pt,parsep=0pt]
\item \textbf{One change per sentence:} use retrieval, the default, which adds 173 prompt tokens per draft and keeps the judgment rate near that of the 10-K writers on the same changes (Section~\ref{sec:res_audit}).
\item \textbf{Paragraphs of several changes or a fixed prompt budget:} use the style or faithful variant, as retrieved sentences lead models to compare changes across firms (Section~\ref{sec:res_variant}).
\item \textbf{Inputs with notes:} the support test keeps almost all of the noted causes, and the cut ones can be listed for the reviewer (Appendix~\ref{app:cause}).
\end{itemize}

\noindent\textbf{Causes required by Item 303.}
Where MD\&A must explain a material change, the slot mode replaces a cut cause with \textit{``[explain driver]''}, so that the preparer supplies the true cause (Appendix~\ref{app:edit}).

\begin{table*}[t]
\centering
\caption{\textbf{Examples of unsupported framing.} Three test changes with the LLM input, the writer's sentence, a plain draft, and the \ours{} output of the same LLM. Row (c) shows a failure of \ours{}.}
\label{tab:case}
\includegraphics[width=\textwidth]{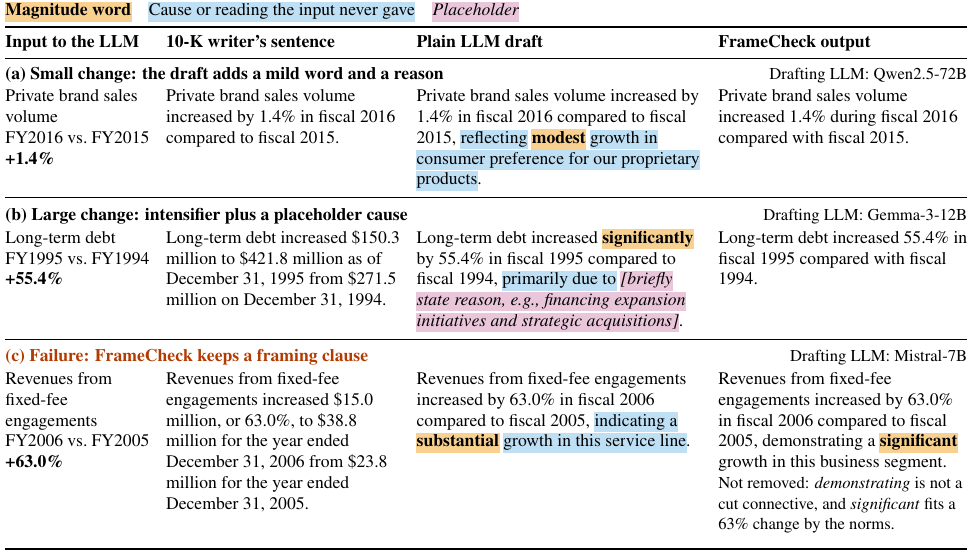}
\end{table*}

\noindent\textbf{Monitoring.}
The judgment rate by size of change, against the writers' rates, serves as a dashboard for a pipeline, and each flag marks a span for review.

\noindent\textbf{Known failure modes.}
The cause check misses causes that start with other verbs or sit in the subject, the word check misses judgments outside its lexicon, and a sentence that reports several changes may be flagged against the wrong size of change, which a reviewer then clears (Appendix~\ref{app:check}).

\section{Examples}
\label{app:examples}

Table~\ref{tab:case} shows three test changes in full. The first two \ours{} outputs lack the judgment word and the unsupported cause or placeholder of the plain draft and keep the number, while the third shows a failure, a judgment set off by a verb outside the connective list that survives the edit.

\end{document}